\documentclass[letterpaper]{article} 
\usepackage[preprint]{aaai2027}  
\usepackage[hyphens]{url}  
\usepackage{graphicx} 
\graphicspath{{figures/}}
\usepackage{natbib}  
\usepackage{caption} 
\usepackage{amsmath}
\usepackage{amsthm}
\usepackage{amssymb}
\usepackage{mathtools}
\usepackage{booktabs}
\usepackage{comment}
\usepackage{graphicx}
\usepackage{diagbox}
\usepackage{multirow}
\usepackage{subcaption}

\theoremstyle{definition}

\title{From Source Bias to Target Evidence: Progressive Context Refinement for Vision-Language Models}
\author{
    Dingyu Zhang\equalcontrib\textsuperscript{\rm 1},
    Jiaxin Qi\equalcontrib\textsuperscript{\rm 2},
    Yan Cui\textsuperscript{\rm 1},
    Jinli Ou\textsuperscript{\rm 1},
    Yuhua Zheng\textsuperscript{\rm 1},
    Jianqiang Huang\textsuperscript{\rm 2}
}

\affiliations{
    \textsuperscript{\rm 1}Hangzhou Institute for Advanced Study,
    University of Chinese Academy of Sciences, Hangzhou, China\\
    \textsuperscript{\rm 2}Computer Network Information Center,
    Chinese Academy of Sciences, Beijing, China\\
    {\small
    \textsuperscript{\rm 1}\texttt{zhangdingyu26@mails.ucas.ac.cn},
    \texttt{cuiyan.ch@gmail.com}\\
    \texttt{oujinli@zuaa.zju.edu.cn},
    \texttt{zhengyuhua@ucas.ac.cn}\\
    \textsuperscript{\rm 2}\texttt{jxqi@cnic.cn},
    \texttt{jqhuang@cnic.cn}
    }
}

\begin{document}

\maketitle

\begin{abstract}

Large Vision-Language Models (LVLMs) have demonstrated strong reasoning capabilities through In-Context Learning (ICL), performing novel tasks via few-shot demonstrations without parameter updates. 
However, the success of current ICL heavily relies on the independent and identically distributed (IID) assumption, requiring fully annotated support sets that share the same distribution as the test queries. 
In open-world applications, this ideal condition is frequently violated, leading to a critical yet underexplored challenge: the inherent ``imperfection'' of available support data. 
To bridge this gap, we formalize a novel and practical ICL setting termed Hybrid-Context ICL (HC-ICL), which constructs a heterogeneous support set by integrating data from two realistic but suboptimal scenarios:
(1) labeled out-of-distribution support sets, which suffer from severe domain shifts, and (2) unlabeled in-distribution support sets, which contain test samples without ground-truth annotations.
Under this paradigm, we investigate whether and how effectively combining these sources can overcome the limitations of relying on either source alone. Crucially, our analysis reveals fundamental alignments and contrasts between HC-ICL and traditional ICL mechanisms. Building upon these insights, we propose Progressive Arbitration of Contextual Evidence (PACE), a novel framework explicitly designed to guide robust hybrid-context reasoning.
Extensive experiments demonstrate that our method establishes a strong baseline for this challenging setting, paving the way for practical multimodal ICL in the wild.

\end{abstract}

\section{Introduction}

Large Vision-Language Models (LVLMs) have demonstrated strong cross-modal understanding and reasoning capabilities, yet their zero-shot performance often remains limited on previously unseen downstream visual reasoning tasks.
To bridge this capability gap, In-Context Learning (ICL)~\cite{brown2020language} has been widely adopted as a training-free adaptation paradigm. In Visual Question Answering (VQA)~\cite{Antol_2015_ICCV}, for example, ICL conditions the model on a few annotated image-question-answer demonstrations, which are prepended to the target query to guide answer generation. A key factor underlying the effectiveness of conventional ICL is the distributional alignment between the provided demonstrations and the target queries. Under such alignment, the model can exploit recurring patterns in the contextual answer distribution~\cite{olsson2022context}, while also implicitly adjusting its vision-language attention toward task-relevant visual and textual cues~\cite{ou2026mllms,schwettmann2023multimodal}.
For example, when querying the specific attributes of an object in a cluttered scene, providing in-distribution support samples with similar visual styles and aligned questions guides the LVLM to effectively recognize the underlying task type. Such context alignment, predicated on the assumption that support examples are independently and identically distributed (IID), underpins the model's robust reasoning.

\begin{figure}[t] 
    \centering 
    \includegraphics[width=\linewidth]{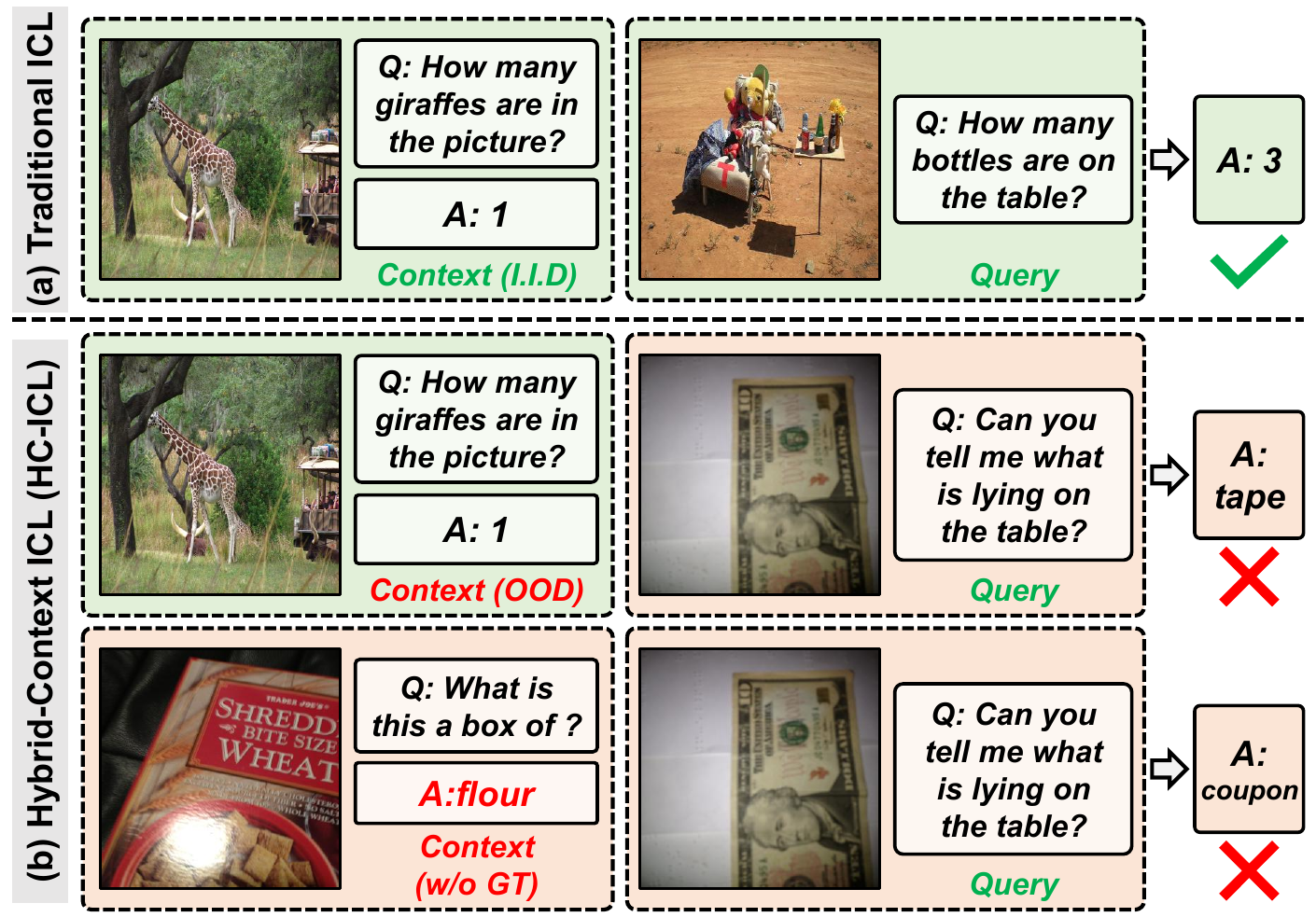}
    \caption{Traditional and Hybrid-Context ICL (background colors denote distinct distributions). (a) Traditional ICL: IID contexts provide correctly aligned reasoning cues. (b) Hybrid-Context ICL: Practical open-world deployments typically face two imperfect context scenarios: labeled OOD samples (which disrupt cross-modal alignment) and unlabeled in-distribution samples paired with model-generated pseudo-labels (which introduce confirmation bias). }
    \label{fig:first_figure}
\end{figure}

However, in open-ended real-world applications, this idealized IID premise is frequently violated, trapping LVLMs in a severe generalization bottleneck during deployment. Specifically, when confronted with unseen target queries, existing source-domain support sets typically exhibit significant distribution shifts, whereas available target-domain samples (which are naturally IID to future queries) inherently lack ground-truth annotations.
This forces the model to make suboptimal choices when constructing imperfect contexts:
(1) Using labeled source-domain pairs as demonstrations provides reliable annotations but introduces substantial distribution shifts relative to the target queries, potentially impairing cross-modal alignment during visual-semantic reasoning.
(2) Using unlabeled target-domain samples as demonstrations preserves input-level distributional alignment but requires model-generated pseudo-labels, making context quality dependent on the model's initial predictions and increasing the risk of error propagation and confirmation bias.
Under either of these degraded conditions, the naive mimicry mechanism underpinning standard ICL inevitably falters due to fundamental feature misalignments and the propagation of erroneous priors. Consequently, a critical research question emerges: when the IID assumption collapses, can ICL still maintain its efficacy under these compromised contexts?


To study this challenge within the standard training-free ICL paradigm, we formalize Hybrid-Context ICL (HC-ICL), a practical setting where demonstrations comprise labeled source-domain samples and pseudo-labeled target-domain samples. Unlike Unsupervised Domain Adaptation~\cite{liu2022deep} and Test-Time Adaptation~\cite{chen2022contrastive}, which often require parameter updates, feature-statistic adaptation, or auxiliary optimization, HC-ICL keeps the LVLM fully frozen and mitigates distribution shift solely through context composition. An effective method should therefore combine the label reliability of source-domain demonstrations with the input alignment of target-domain samples, while suppressing source bias, pseudo-label noise, and visual-reasoning shortcuts.


To meet this desideratum, we propose Progressive Arbitration of Contextual Evidence, a training-free framework that conceptualizes hybrid-context construction as Bayesian-inspired evidence arbitration and progressive posterior refinement. Under a latent-concept view of ICL, the composition of demonstrations determines how contextual evidence shapes the model's implicit posterior over task-relevant concepts. Our contrastive mechanism distinguishes target-conditioned predictive evidence from the nuisance effects induced by target-image-independent shortcuts and source-context bias, prioritizing candidates whose predictive contributions extend beyond both effects. Building on this refined evidence basis, a reliability-aware progressive evolution process gradually incorporates consistent target-domain pseudo-labels into a dynamic pool, steering the contextual posterior toward target-compatible concepts while containing error propagation. Experiments across five VQA datasets demonstrate consistent robustness gains under imperfect contexts.

In summary, the contributions of this work are threefold:
\begin{itemize}
    \item We highlight that conventional ICL heavily relies on the idealized IID assumption, and we systematically formalize two prevalent but imperfect context sets in real-world deployments: labeled out-of-distribution sets with severe domain shifts, and unlabeled target-domain sets relying on noisy pseudo-labels.
    
    \item We propose a highly challenging Hybrid-Context ICL (HC-ICL) setting that integrates the aforementioned imperfect cases. Through extensive empirical studies, we reveal novel phenomena within this setting that are distinctly different from traditional ICL mechanisms.
    
    \item Based on our findings, we propose Progressive Arbitration of Contextual Evidence. By using a log-predictive evidence margin to suppress target-image-independent and source-context-induced nuisance predictions, and by progressively accumulating reliable target-domain evidence through a dynamic pool, our method establishes a strong and robust baseline for HC-ICL.
\end{itemize}
\section{Related Work}

\subsection{In-Context Learning}

In-context learning (ICL) adapts language models to downstream tasks using a few demonstrations \cite{brown2020language,wei2022chain} and has been extended to multimodal models \cite{awadalla2023openflamingo}, with VQA as a common benchmark \cite{peng2024live,dang2024sadl}. Demonstration selection is crucial: random sampling is unstable \cite{zhao2021calibrate}, while similarity-based retrieval \cite{liu2022makes,li2024configure} may introduce spurious correlations and reasoning shortcuts \cite{zhou2024explore,baldassini2024makes}. Recent approaches use reinforcement learning \cite{zhang2022active,laskin2022context}, contrastive learning \cite{mo2024c,gao2024customizing}, or information theory \cite{nanda2025insquad,wu2023self}, but generally assume distributionally aligned demonstrations and queries.

OOD studies examine generalization beyond pretraining functions or algorithms \cite{goddard2025can,garg2022can,wang2024can}, as well as shifts in geometric structures and symbols \cite{kwon2025out,song2025out}. However, most evaluations rely on synthetic data or unimodal language tasks, leaving cross-domain demonstration selection in realistic multimodal settings underexplored.

\subsection{Large Vision-Language Models}

Early LVLMs such as Flamingo \cite{alayrac2022flamingo} and BLIP-2 \cite{li2023blip} established multimodal reasoning by connecting language models with visual encoders. Subsequent adaptation follows two main directions: instruction tuning for zero-shot task following, as in LLaVA \cite{liu2023visual} and Qwen2-VL \cite{wang2024qwen2}, and multimodal ICL using interleaved image-text demonstrations. OpenFlamingo \cite{awadalla2023openflamingo} provides a representative platform for studying multimodal ICL. Despite related mechanisms across LVLM architectures \cite{baldassini2024makes,doveh2024towards}, their robustness under severe domain shifts remains insufficiently understood.
\section{Problem Formulation}
\label{section3}

\begin{figure*}[t] 
    \centering
    \includegraphics[width=\textwidth]{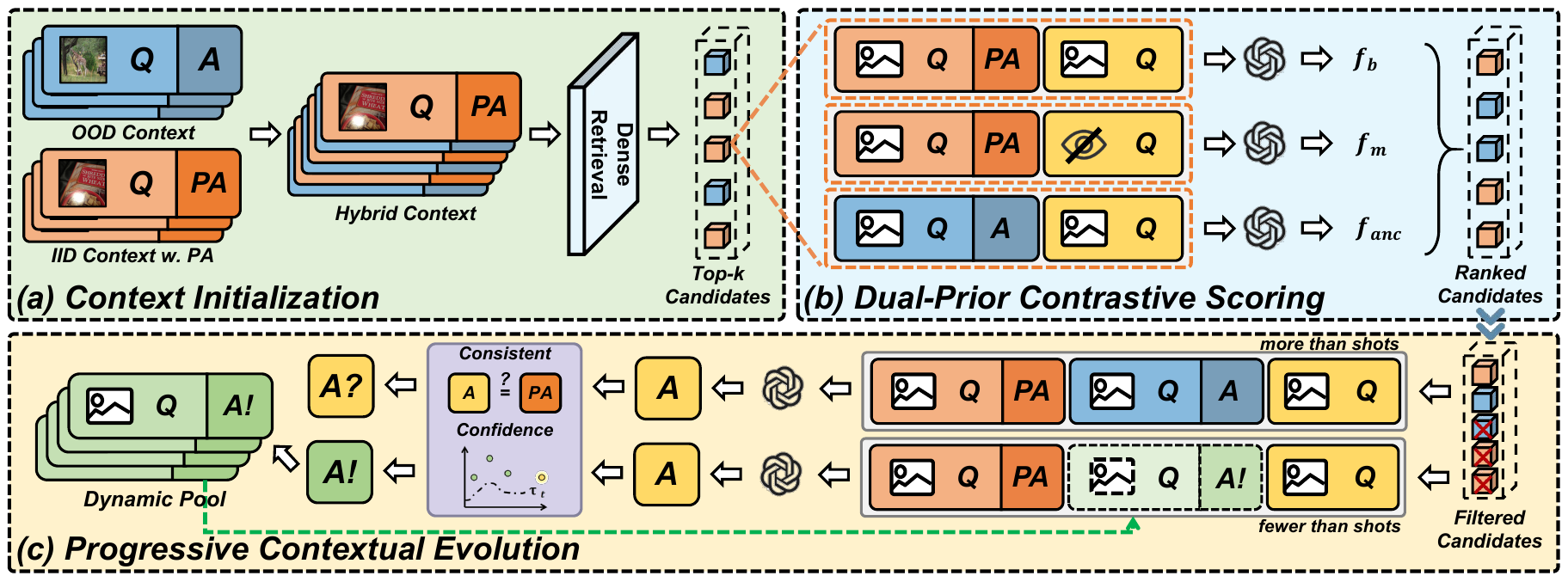}
    
    \caption{Overview of the proposed Progressive Arbitration of Contextual Evidence framework. By strictly decoupling spurious textual shortcuts via contrastive scoring (b) and safely accumulating target-domain pseudo-labels through dynamic evolution (c), this training-free pipeline restores robust cross-modal reasoning under severe domain shifts.}
    
    \label{fig:method_overview}
\end{figure*}

In this section, we formulate the Hybrid-Context In-Context Learning (HC-ICL) paradigm. We first revisit the standard ICL formulation and its strict data assumptions. Next, we formalize our novel hybrid setting, detailing the construction of imperfect support sets in open-world scenarios and the challenges from cross-domain shifts and pseudo-label noise.

\subsection{Traditional In-Context Learning}
\label{3.1}
In-context learning (ICL) enables LVLMs to adapt to unseen tasks via a few-shot demonstration context $C$, without requiring parameter updates. While broadly applicable, this paper focuses on the highly challenging multimodal reasoning scenario, rigorously formalizing it within the Visual Question Answering (VQA) framework.

Specifically, we define a standard data instance in the VQA data space as a tuple $(v,q,a)$, comprising the visual input, textual query, and ground-truth answer. Let $\mathcal{M}$ denote a base LVLM. Given a test input $(v_{t},q_{t})$, $\mathcal{M}$ autoregressively decodes the predicted answer $\hat{a}_{t}$ by maximizing the likelihood of the generation sequence under the zero-shot setting:

\begin{equation}
    \hat{a}_{t} = \mathop{\arg\max}_{a} P_{\mathcal{M}}(a|v_{t},q_{t}).
\end{equation}

Unlike zero-shot inference, standard ICL utilizes external demonstrations to guide the model's reasoning. Given a support set $S=\{s_i\}_{i=1}^{N}$ containing $N$ labeled instances where $s_i=(v_i,q_i,a_i)$, ICL employs a selection function $\mathcal{F}_s(v_t,q_t,S)$ to retrieve $K$ samples. The selected demonstrations are concatenated to form the context
$C=[s_{k_1};s_{k_2};\dots;s_{k_K}]$, where
$s_{k_j}=\mathcal{F}_s(v_t,q_t,S)_j$ denotes the $j$-th retrieved sample and $[;]$ denotes concatenation. The ICL decoding process is then formulated as:
\begin{equation}
   \hat{a}_t
   =
   \operatorname*{arg\,max}_{a}
   P_{\mathcal{M}}(a \mid C, v_t, q_t).
   \label{eq:icl}
\end{equation}
In practice, the selection function $\mathcal{F}_s$ is typically instantiated via similarity-based retrieval in a specific feature space (e.g., visual, textual, or cross-modal \cite{gao2021making}) or through random sampling.

Theoretically, the efficacy of standard ICL is predicated upon a strict IID assumption. Specifically, it assumes a perfect distributional alignment where both the selected supports from $S$ and the unseen test instance are drawn from the same target distribution $\mathcal{D}_t$:
\begin{equation}
    s_i \overset{\text{i.i.d.}}{\sim} \mathcal{D}_t, \;\forall s_i \in S \; \text{and} \; (v_t, q_t, a_t) \sim \mathcal{D}_t.
\end{equation}
Under this ideal condition, the retrieved context $C$ serves as an unbiased empirical approximation of $\mathcal{D}_t$. By demonstrating exact, in-distribution ground-truth mappings, $C$ implicitly calibrates the cross-modal attention of $\mathcal{M}$, facilitating robust few-shot adaptation.

\subsection{Hybrid-Context In-Context Learning}
\label{3.2}
In open-world deployments, the idealized IID assumption frequently collapses. In practice, obtaining massive manually annotated data that perfectly aligns with the target distribution $\mathcal{D}_{t}$ is prohibitively expensive \cite{ganin2016domain}. Instead, we formalize a more realistic {Hybrid-Context ICL (HC-ICL)} setting, where the context is constructed from two accessible but imperfect sources:

\noindent\textit{(1) Labeled Out-of-Distribution Data}. A support set $S_{l}=\{s_{i}\}_{i=1}^{N_{l}}$ consisting of fully annotated instances from a source domain. While these instances provide high-quality ground-truth mappings, their underlying joint distribution diverges from the target domain:
\begin{equation}
    s_{i} \sim \mathcal{D}_{l},  \;\forall s_{i} \in S_l, \; \text{where} \; \mathcal{D}_{l} \neq \mathcal{D}_{t}.
\end{equation}
Although the labels are accurate, their visual and linguistic features may deviate from the target query $(v_{t}, q_t)$, inevitably introducing domain biases and potential negative transfer.




\noindent\textit{(2) Unlabeled Target Domain Data}. A support set ${S}_u=\{(v_i,q_i)\}_{i=1}^{N_u}$ consisting of unlabeled instances from the target domain is available. To use these samples as demonstrations, we generate pseudo-labels and obtain the pseudo-labeled support set $\hat{S}_u=\{\hat{s}_i\}_{i=1}^{N_u}$, while their assigned pseudo-labels lack ground-truth verification:

\begin{equation}
    \hat{s}_i = (v_i, q_i, \hat{a}_i)\sim \mathcal{D}_u, \;\forall \hat{s}_i \in \hat{S}_u, \; \text{where}\; \mathcal{D}_u \neq \mathcal{D}_t,
\end{equation}
where $\hat{a}_i$ denotes the pseudo-label generated by the model based on the input $(v_i, q_i)$. Although $\mathcal{D}_u$ and $\mathcal{D}_t$ share an identical marginal distribution over the inputs $(v,q)$, their joint distributions differ because the conditional probability of the pseudo-labels deviates from the ground truth. This discrepancy inevitably introduces noise and confirmation bias into the demonstration context.

Formally, we define Hybrid-Context ICL as a setting where the available
support set $S_h$ is a heterogeneous union of these two sources, i.e.,
$S_h = S_l \cup \hat{S}_u$, addressing realistic imperfect support
conditions overlooked by traditional ICL.

\section{Our Method}

\begin{table*}[t]
\centering

\scriptsize
\renewcommand{\arraystretch}{1.08}
\setlength{\tabcolsep}{1.5pt}

\begin{minipage}[t]{0.49\textwidth}
\vspace{0pt}
\centering
\textbf{Source = VQAv2}\\[1mm]
\begin{tabular*}{\linewidth}{@{\extracolsep{\fill}}lcccccccc@{}}
\toprule
\textbf{Target}
& \textbf{Base}
& \textbf{TT-RAA}
& \textbf{U-ICL}
& \textbf{BCA}
& \textbf{MMICES}
& \textbf{RICES}
& \textbf{DARA}
& \textbf{Ours}
\\
\midrule
\textbf{VQAv2}
& 47.87 & 48.55 & 44.36 & 48.67 & 48.77 & 48.56 & 47.41 & \textbf{49.94} \\
\textbf{Number}
& 37.02 & 26.53 & 38.21 & 36.95 & 38.37 & 39.93 & \textbf{40.58} & 39.79 \\
\textbf{VizWiz}
& \textbf{19.96} & 18.12 & 18.53 & 19.34 & 19.35 & 19.44 & 17.65 & 19.85 \\
\textbf{OK-VQA}
& 28.14 & 29.07 & 28.49 & 29.30 & 29.72 & 29.34 & 28.83 & \textbf{30.26} \\
\textbf{PathVQA}
& 20.08 & 14.73 & 18.59 & 20.05 & 19.19 & 21.66 & 19.86 & \textbf{23.89} \\
\midrule
\textbf{Avg.}
& 30.61 & 27.40 & 29.64 & 30.86 & 31.08 & 31.79 & 30.87 & \textbf{32.75} \\
\bottomrule
\end{tabular*}
\end{minipage}
\hfill
\begin{minipage}[t]{0.49\textwidth}
\vspace*{0.35mm}  
\centering
\textbf{Source = VizWiz}\\[1mm]
\begin{tabular*}{\linewidth}{@{\extracolsep{\fill}}lcccccccc@{}}
\toprule
\textbf{Target}
& \textbf{Base}
& \textbf{TT-RAA}
& \textbf{U-ICL}
& \textbf{BCA}
& \textbf{MMICES}
& \textbf{RICES}
& \textbf{DARA}
& \textbf{Ours}
\\
\midrule
\textbf{VQAv2}
& 31.47 & 20.65 & 38.72 & 32.78 & 34.35 & 37.30 & 35.08 & \textbf{42.20} \\
\textbf{Number}
& 33.43 & 18.20 & 34.90 & 31.98 & 33.67 & 35.54 & 41.45 & \textbf{43.00} \\
\textbf{VizWiz}
& 36.98 & 38.52 & 40.76 & 39.92 & 40.92 & 42.56 & \textbf{49.54} & 49.18 \\
\textbf{OK-VQA}
& 20.53 & 17.36 & 24.08 & 19.47 & 22.90 & \textbf{24.58} & 20.88 & 22.76 \\
\textbf{PathVQA}
& 8.95 & 5.48 & 10.56 & 9.00 & 9.36 & 12.51 & 9.76 & \textbf{19.80} \\
\midrule
\textbf{Avg.}
& 26.27 & 20.04 & 29.80 & 26.63 & 28.24 & 30.50 & 31.34 & \textbf{35.39} \\
\bottomrule
\end{tabular*}
\end{minipage}

\vspace{2mm}

\begin{minipage}[t]{0.49\textwidth}
\vspace{0pt}
\centering
\textbf{Source = OK-VQA}\\[1mm]
\begin{tabular*}{\linewidth}{@{\extracolsep{\fill}}lcccccccc@{}}
\toprule
\textbf{Target}
& \textbf{Base}
& \textbf{TT-RAA}
& \textbf{U-ICL}
& \textbf{BCA}
& \textbf{MMICES}
& \textbf{RICES}
& \textbf{DARA}
& \textbf{Ours}
\\
\midrule
\textbf{VQAv2}
& 35.79 & 29.55 & 38.20 & 38.70 & 38.16 & 39.85 & 40.62 & \textbf{42.25} \\
\textbf{Number}
& 36.70 & 13.10 & 35.07 & 39.85 & 38.95 & 39.85 & \textbf{43.74} & 43.58 \\
\textbf{VizWiz}
& \textbf{19.92} & 16.05 & 15.24 & 18.18 & 18.58 & 19.36 & 15.99 & 18.38 \\
\textbf{OK-VQA}
& 35.74 & 35.79 & 35.88 & \textbf{36.70} & 35.03 & 36.68 & 32.34 & 36.60 \\
\textbf{PathVQA}
& 15.67 & 10.94 & 12.96 & 15.37 & 13.37 & 17.64 & 16.78 & \textbf{21.57} \\
\midrule
\textbf{Avg.}
& 28.76 & 21.09 & 27.47 & 29.76 & 28.82 & 30.68 & 29.89 & \textbf{32.48} \\
\bottomrule
\end{tabular*}
\end{minipage}
\hfill
\begin{minipage}[t]{0.49\textwidth}
\vspace{0pt}
\centering
\textbf{Source = PathVQA}\\[1mm]
\begin{tabular*}{\linewidth}{@{\extracolsep{\fill}}lcccccccc@{}}
\toprule
\textbf{Target}
& \textbf{Base}
& \textbf{TT-RAA}
& \textbf{U-ICL}
& \textbf{BCA}
& \textbf{MMICES}
& \textbf{RICES}
& \textbf{DARA}
& \textbf{Ours}
\\
\midrule
\textbf{VQAv2}
& 26.41 & 15.14 & 30.18 & 27.91 & 26.22 & 28.70 & 29.38 & \textbf{36.55} \\
\textbf{Number}
& 33.76 & 6.46 & 36.67 & 36.23 & 35.35 & 38.11 & \textbf{41.23} & 39.00 \\
\textbf{VizWiz}
& 11.38 & 7.94 & 7.89 & 10.77 & 8.88 & 10.77 & 7.96 & \textbf{12.83} \\
\textbf{OK-VQA}
& 19.51 & 14.82 & 22.74 & 22.04 & 20.87 & 23.01 & \textbf{25.30} & 24.61 \\
\textbf{PathVQA}
& 37.10 & 34.45 & 35.13 & 36.94 & \textbf{38.94} & 36.94 & 27.96 & 38.90 \\
\midrule
\textbf{Avg.}
& 25.63 & 15.76 & 26.52 & 26.78 & 26.05 & 27.51 & 26.37 & \textbf{30.38} \\
\bottomrule
\end{tabular*}
\end{minipage}

\caption{Source-target evaluation of 4-shot ICL under hybrid-context construction (Accuracy, \%), covering both cross-domain and same-dataset settings. Results with Number as the source domain are reported in the Supplementary Material. Base denotes the initial OOD-ICL predictions used as pseudo-labels, while TT-RAA, U-ICL, BCA, MMICES, RICES, and DARA are competing methods. Across all reported source domains, our method achieves the highest source-wise average accuracy. Best results for each target under each source domain and for each source-domain average are in bold.}
\label{tab:main_results}

\end{table*}

\subsection{Bayesian View of Hybrid Evidence} 

Recent studies interpret ICL as implicit Bayesian inference over a latent concept $z$ shared by demonstrations \cite{xie2021explanation,zhang2023and,panwar2024context}. Given a target query $x_t=(v_t,q_t)$ and context $C$, the prediction of the frozen LVLM $\mathcal{M}$ can be viewed as an approximate posterior predictive distribution: \begin{equation} P_{\mathcal{M}}(a_t\mid x_t,C) \approx \int P(a_t\mid x_t,z)\,p(z\mid C)\,dz. \label{eq:bayesian_view} \end{equation} 
Under standard ICL, in-distribution demonstrations provide clean evidence
for shaping the implicit posterior $p(z\mid C)$. In HC-ICL, however, the
evidence is inherently imperfect: labeled source-domain demonstrations
offer reliable answer mappings but may bias the posterior toward source
concepts, while pseudo-labeled target demonstrations are input-aligned
but may inject erroneous answer evidence. Therefore, our objective is not
merely to retrieve similar demonstrations, but to select evidence that
shifts the implicit posterior toward target-compatible concepts while
suppressing nuisance predictive tendencies induced by both imperfect
sources.

\subsection{Dual-Prior Contrastive Scoring (DPCS)}

As illustrated in Figure~\ref{fig:method_overview}(a)--(b),for each target query $(v_t,q_t)$, we first retrieve the top
$K_{\mathrm{recall}}$ candidates from the hybrid support pool
$\mathcal{S}_h$ using similarity-based retrieval. This stage serves
only as coarse recall, since high surface similarity may reinforce
shortcut predictions under domain shift. We therefore evaluate whether
each candidate $e$ provides predictive support beyond two controlled
reference conditions.

For each candidate $e$, we identify the highest-probability answer token
under the complete candidate-target context:
\begin{equation}
a^*
=
\arg\max_a
P_{\mathcal{M}}(a\mid e,v_t,q_t).
\label{eq:primary_token}
\end{equation}
Scores are computed from the log-softmax distribution at the first
decoding step. The full-context log-predictive score is
\begin{equation}
f_b
=
\log
P_{\mathcal{M}}(a^*\mid e,v_t,q_t),
\label{eq:full_score}
\end{equation}
which measures the normalized predictive support for $a^*$ when both
the candidate and target image are available.

Here, \emph{prior} denotes an operational predictive tendency induced
by the frozen LVLM under a controlled context condition and used for
comparative candidate scoring, rather than a formal Bayesian prior over
the latent concept $z$.

\paragraph{Target-Image-Independent Prior Score ($f_m$).}
We replace the target image $v_t$ with a fixed uninformative placeholder
$v_m$ and evaluate the same answer token:
\begin{equation}
f_m
=
\log
P_{\mathcal{M}}(a^*\mid e,v_m,q_t).
\label{eq:mask_score}
\end{equation}
By holding $a^*$ fixed while removing the target image, $f_m$ captures
predictive support arising from contextual regularities, recurring
question patterns, or answer-frequency shortcuts rather than target
visual evidence.

\paragraph{Source-Context Prior Score ($f_{\mathrm{anc}}$).}
We replace the candidate with a representative source-domain anchor:
\begin{equation}
f_{\mathrm{anc}}
=
\log
P_{\mathcal{M}}(a^*\mid e_{\mathrm{anc}},v_t,q_t).
\label{eq:anchor_score}
\end{equation}
To construct $e_{\mathrm{anc}}$, we compute the centroid of the
normalized joint visual-linguistic representations of all labeled
source samples and select the closest sample as the anchor. Thus,
$f_{\mathrm{anc}}$ measures predictive support induced by a generic
source-domain context rather than the specific candidate $e$.

\paragraph{Dual-Prior Predictive Evidence Margin.}
Using the three log-predictive scores, we define
\begin{equation}
\begin{aligned}
\mathcal{S}_{\mathrm{DPCS}}(e\mid v_t,q_t)
&=
\log
\frac{\exp(f_b)}
{\max\!\left(\exp(f_m),\exp(f_{\mathrm{anc}})\right)}
\\
&=
f_b-\max(f_m,f_{\mathrm{anc}}).
\end{aligned}
\label{eq:dpcs}
\end{equation}
The score measures the log-predictive advantage of the complete
candidate context over the stronger operational prior. Although it is
not a formal Bayes factor, it provides a Bayesian-inspired comparison
of predictive evidence under competing context conditions. We retain
candidates satisfying $\mathcal{S}_{\mathrm{DPCS}}>0$ and rank them in
descending order.

\subsection{Progressive Contextual Evolution}

As illustrated in Figure~\ref{fig:method_overview}(c), the contrastive filter may return fewer candidates than the required context length. During cold start, the remaining positions are filled with randomly sampled source-domain demonstrations. Although these fallback samples may be domain-mismatched, they provide reliable annotations and serve as a conservative context prior when trustworthy target-domain evidence is still insufficient.

As inference proceeds, reliable pseudo-labeled target samples are accumulated in a dynamic pool $\mathcal{S}_{\mathrm{dyn}}$ and progressively replace source-domain fallback samples. From a Bayesian view, this process sequentially refines the evidence conditioning the implicit posterior over $z$: early contexts rely more on source-domain observations, while later contexts contain more target-aligned evidence.

Formally, for a target sample $x_i=(v_i,q_i)$, let
$\hat{a}_i^{(0)}$ denote its initial pseudo-label,
$\hat{a}_i^{(C_i)}$ denote its prediction under the current context $C_i$,
and $\gamma_i$ denote the corresponding generation confidence.
We define a reliability gate:
\begin{equation}
r_i
=
\mathbb{I}\!\left[
\hat{a}_i^{(0)}=\hat{a}_i^{(C_i)}
\right]
\cdot
\mathbb{I}\!\left[
\gamma_i\ge\tau_i
\right].
\label{eq:reliability_gate}
\end{equation}
Accordingly, the dynamic pool is defined as
$\mathcal{S}_{\mathrm{dyn}}^{(t)}
=
\{(x_i,\hat{a}_i^{(C_i)})
\mid
i<t,\,
\hat{a}_i^{(0)}=\hat{a}_i^{(C_i)},\,
\gamma_i\ge\tau_i\}$.
The threshold is updated as
$\tau_i=\max(\tau_{\min},\mu_W-\sigma_W)$,
where $\mu_W$ and $\sigma_W$ denote the mean and standard deviation of
the confidence scores of the most recent $W$ admitted samples.

Conceptually, this dynamic pool induces a sequential posterior update. To avoid repeatedly counting previously admitted samples, we update the implicit posterior only with newly admitted target evidence:
\begin{equation}
p_{t+1}(z)
\propto
p_t(z)
\prod_{(x_i,\hat{a}_i)\in
\Delta \mathcal{S}_{\mathrm{dyn}}^{(t+1)}}
P_{\mathcal{M}}(\hat{a}_i\mid x_i,z),
\label{eq:posterior_update}
\end{equation}
where $\Delta \mathcal{S}_{\mathrm{dyn}}^{(t+1)}
=
\mathcal{S}_{\mathrm{dyn}}^{(t+1)}
\setminus
\mathcal{S}_{\mathrm{dyn}}^{(t)}$ denotes the newly admitted samples, and $p_t(z)$ denotes the implicit posterior induced by the available context at step $t$. We emphasize that Equation~\ref{eq:posterior_update} is an interpretive formulation rather than an explicit parameter update: the frozen LVLM is unchanged, and the posterior refinement is realized by updating the evidence pool used for subsequent ICL contexts.

In this way, Progressive Contextual Evolution gradually shifts the context from source-biased but label-reliable evidence toward target-compatible evidence, while the reliability gate prevents low-confidence pseudo-labels from accumulating and amplifying confirmation bias.

\section{Experiments}

\subsection{Datasets and Evaluation Metrics}

\paragraph{Datasets and Cross-Domain Setup.}
We evaluate our framework on five VQA datasets covering diverse distribution shifts: \textbf{VQAv2} \cite{Goyal_2017_CVPR} serves as the natural-image baseline; \textbf{OK-VQA} \cite{marino2019ok} requires external knowledge; \textbf{VizWiz} \cite{Gurari_2018_CVPR} contains low-quality images captured by visually impaired users; \textbf{PathVQA} \cite{he2020pathvqa} focuses on medical pathology images; and the \textbf{COCOQA Number} subset \cite{ren2015exploring} targets counting questions.

We evaluate every ordered dataset pair $(A,B)$, including cross-domain
$(A\neq B)$ and same-dataset $(A=B)$ settings. The training split of $A$
forms the labeled source pool $S_l$, while the designated validation/test
split of $B$ is used transductively as both the unlabeled target pool $S_u$
and the evaluation set, with annotations withheld during inference. All target
inputs are accessible, but each query is excluded from its own retrieval
candidates and dynamic pool to avoid trivial self-retrieval. Using $S_l$ as
demonstrations, we apply Equation~\ref{eq:icl} to pseudo-label each
$(v_t,q_t)\in S_u$, yielding $\hat{S}_u$, and construct the hybrid pool
$S_h=S_l\cup\hat{S}_u$.

\paragraph{Evaluation Metrics.}
We apply standard VQA answer normalization. For VQAv2, OK-VQA, and VizWiz, which provide multiple human annotations, we use soft-voting accuracy:
$\min(1,\text{match\_count}/3)$. For the single-annotation COCOQA Number and PathVQA datasets, we use Exact Match.

\subsection{Implementation Details}

\paragraph{Model and Inference Configurations.}
We use OpenFlamingo \cite{awadalla2023openflamingo} with the MPT-7B language model \cite{mosaicml2023introducing} and CLIP ViT-L/14 visual encoder \cite{radford2021learning}. Inputs follow the standard interleaved image-text format, and generation is limited to 10 tokens. We use the maximum softmax probability at the first decoding step as the generation confidence $\gamma_i$ for dynamic selection. 

\paragraph{Baselines.}
We compare Progressive Arbitration of Contextual Evidence with RICES, MMICES \cite{chen2025can}, DARA \cite{chen2025true}, U-ICL \cite{gadetsky2025large}, BCA \cite{zhou2025bayesian}, and TT-RAA \cite{fan2025test}. For RICES, we evaluate three retrieval variants based on image, question, and joint question-answer similarity, and report their maximum score in each source-target setting to represent the best performance among similarity-based retrieval methods.

\paragraph{Hyperparameter Settings.}
All methods use $N_{\mathrm{shots}}=4$. For our method, we set $K_{\mathrm{recall}}=15$, $W=50$, and $\tau_{\min}=0.1$. All reported results use a random seed of 42. Three-seed evaluations on representative settings yield an average standard deviation of only 0.02 percentage points. Additional stability, backbone, and shot-number results are provided in Supplementary.

\begin{figure}[t]
    \centering
    \includegraphics[width=\linewidth]{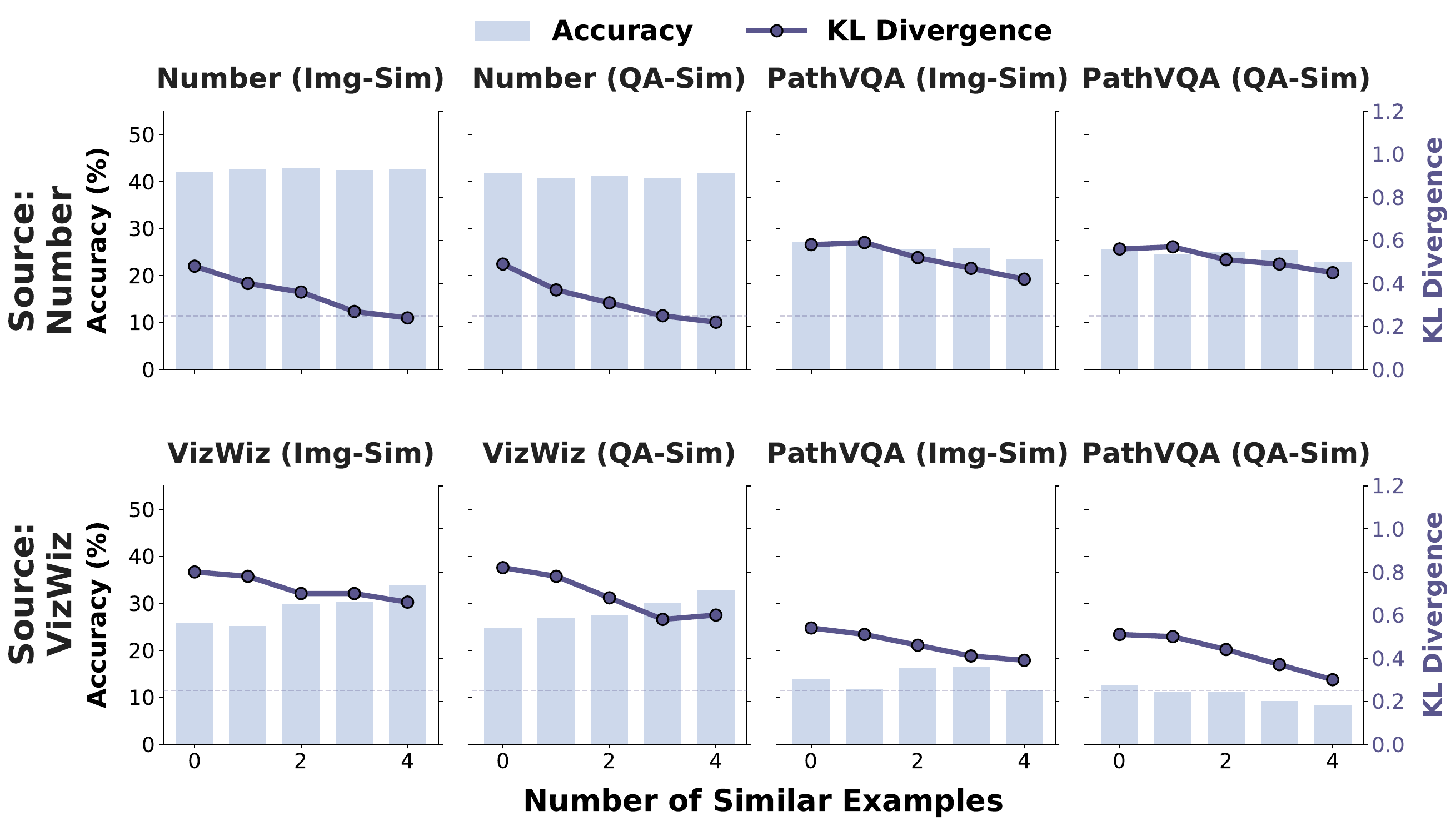}
    \caption{Visual reliance and accuracy under different numbers of similarity-retrieved demonstrations in in-domain and cross-domain settings. Bars show accuracy, while lines show KL divergence as a measure of visual reliance.}
    \label{fig:kl_all}
\end{figure}

\subsection{Main Results}

Table~\ref{tab:main_results} reports four source domains, with Number-source
results provided in Supplementary. Unless noted otherwise, we discuss the
complete $5\times5$ matrix. Our method achieves the highest target-averaged
accuracy for every source and reaches 32.61\% over all 25 pairs,
outperforming RICES by 1.97 points (32.61\% vs.\ 30.64\%).

\paragraph{Robustness under Severe Domain Shifts.}
Our advantage is more pronounced under large source--target gaps. With
VizWiz and PathVQA as source domains, our method improves the strongest
baseline by 4.05 and 2.87 points, respectively. It further raises
VizWiz-to-PathVQA accuracy from 12.51\% to 19.80\% and
PathVQA-to-VQAv2 from 30.18\% to 36.55\%. These results indicate that
our selection mechanism mitigates the effects of domain-mismatched
demonstrations and unreliable pseudo-labels.

\paragraph{Stability in In-Domain Settings.}
Our method also remains competitive when the source and target data come from 
the same dataset, as shown in Table~\ref{tab:main_results} and the supplementary 
material. It achieves the best results on VQAv2 and Number, while trailing the 
strongest baseline by only 0.10, 0.36, and 0.04 points on OK-VQA, VizWiz, and 
PathVQA, respectively. These results indicate improved cross-domain robustness 
without sacrificing same-dataset performance.

\paragraph{Ablation Study.}
We conduct a stage-wise ablation by partitioning queries according to
whether DPCS retains sufficient candidates. For queries requiring no
supplementation, DPCS alone achieves 33.98\% accuracy, outperforming the
corresponding retrieval baseline by 3.39 percentage points. For queries
with insufficient retained candidates, progressively supplementing the
incomplete context with dynamic-pool evidence improves accuracy from
28.16\% to 32.18\% $(+4.02$ points$)$. Moreover, 73.81\% of the
pseudo-labels admitted to the dynamic pool are correct according to the
withheld ground-truth annotations, demonstrating the reliability of the
admitted target-domain evidence. These results verify the respective
contributions of dual-prior filtering and progressive contextual
evolution. Detailed subset and dynamic-pool statistics are provided in
the Supplementary.

\paragraph{Efficiency.}
Dual-prior scoring introduces additional inference cost compared with
lightweight similarity-based retrieval. Nevertheless, the overall runtime
is only approximately $1.1\times$--$1.3\times$ that of the evaluated
baseline methods and remains comparable to other competitive approaches.
Runtime comparisons are provided in Supplementary.

\begin{table}[t]
    \centering

    \renewcommand{\arraystretch}{1.28}

    \resizebox{\linewidth}{!}{%
        \begin{tabular}{l|ccccc|c|c}
            \toprule
            \multirow{2}{*}{
                \diagbox[width=7em]
                {\textbf{Test}}
                {\textbf{Support}}
            }
            & \multicolumn{5}{c|}{\textbf{OOD}}
            & \multirow{2}{*}{\textbf{Zero-shot}}
            & \multirow{2}{*}{\textbf{U-ICL}}
            \\
            \cmidrule(lr){2-6}
            & \textbf{VQAv2}
            & \textbf{Number}
            & \textbf{VizWiz}
            & \textbf{OK-VQA}
            & \textbf{PathVQA}
            & {}
            & {}
            \\
            \midrule

            \textbf{VQAv2}
            & -
            & 42.22
            & 31.47
            & 35.79
            & 26.41
            & 14.96
            & 42.52
            \\

            \textbf{Number}
            & 37.02
            & -
            & 33.43
            & 36.70
            & 33.76
            & 22.36
            & 40.68
            \\

            \textbf{VizWiz}
            & 19.96
            & 18.10
            & -
            & 19.92
            & 11.38
            & 8.39
            & 28.12
            \\

            \textbf{OK-VQA}
            & 28.14
            & 24.40
            & 20.53
            & -
            & 19.51
            & 11.68
            & 32.37
            \\

            \textbf{PathVQA}
            & 20.08
            & 27.02
            & 8.95
            & 15.67
            & -
            & 10.21
            & 28.42
            \\

            \bottomrule
        \end{tabular}%
    }
    \caption{Accuracy (\%) comparison of Zero-shot, OOD ICL, and Unsupervised ICL (U-ICL) across five target domains, with ICL contexts retrieved by question similarity.}

    \label{tab:ood_matrix_q}
\end{table}

\subsection{Analysis and Discussion}

We first quantify the limitations of standard ICL under OOD contexts. We then analyze how context composition affects visual reliance and structural diversity, revealing two factors that account for the degradation of similarity-based ICL under domain shifts.

\noindent\textbf{Empirical Evidence of the OOD Dilemma.}
For each target dataset, we compare Zero-shot inference, OOD ICL using
labeled source-domain demonstrations, and Unsupervised ICL (U-ICL)
\cite{gadetsky2025large} using pseudo-labeled target-domain contexts
under the same evaluation protocol.

As shown in Table~\ref{tab:ood_matrix_q}, Zero-shot, OOD ICL,
and U-ICL achieve average accuracies of 13.52\%, 25.52\%, and
34.42\%, respectively; even selecting the best OOD source for
each target reaches only 30.87\%. These results show that reliable
but distribution-shifted labels are less effective than target-aligned
pseudo-labeled contexts, motivating their complementary use in HC-ICL.

\begin{table}[t]
    \centering

    \footnotesize
    \renewcommand{\arraystretch}{1.12}
    \setlength{\tabcolsep}{3pt}

    \begin{tabular}{@{}lccccc@{}}
        \toprule
        \multirow{2}{*}{\textbf{Test Set}}
        & \multicolumn{3}{c}{\textbf{Accuracy (\%)}}
        & \multicolumn{2}{c}{
            $\boldsymbol{D_{\mathrm{KL}}(\mathrm{Clean}\Vert\mathrm{Blur})}$
        }
        \\
        \cmidrule(lr){2-4}
        \cmidrule(lr){5-6}
        & \textbf{Clean}
        & \textbf{Mid Blur}
        & \textbf{High Blur}
        & \textbf{Mid}
        & \textbf{High}
        \\
        \midrule
        OK-VQA
        & 14.36
        & 12.17
        & 9.54
        & 0.1216
        & 0.1637
        \\
        Number
        & 24.52
        & 22.03
        & 18.81
        & 0.2715
        & 0.3202
        \\
        \bottomrule
    \end{tabular}
    \caption{Validation of $D_{\mathrm{KL}}$ with OOD VizWiz support. Random context selection isolates visual degradation from semantic similarity; Mid and High Blur denote two levels of Gaussian perturbation.}
    \label{tab:kl_evidence}
\end{table}

This trade-off suggests that context quality depends not only on label reliability but also on how demonstrations influence the model's use of visual and textual information. We investigate these effects through the following analyses.

\noindent\textbf{Similarity Retrieval Reduces Visual Reliance.}
We measure the model's visual reliance $\Phi(v_t;C)$ using the KL divergence between the output distributions produced from the original target image $v_t$ and a degraded counterpart $v_{\mathrm{deg}}$:
\[
\Phi(v_t;C)=D_{\mathrm{KL}}\left(
P_{\mathcal{M}}(\cdot\mid C,v_t,q_t)
\parallel
P_{\mathcal{M}}(\cdot\mid C,v_{\mathrm{deg}},q_t)
\right).
\]
A larger divergence indicates greater sensitivity to visual evidence, whereas a value near zero suggests limited reliance on the target image. We assess this metric through progressive Gaussian degradation under an OOD setting, using randomly selected contexts to exclude the influence of semantic similarity. As shown in Table~\ref{tab:kl_evidence}, stronger visual degradation consistently reduces accuracy while increasing $D_{\mathrm{KL}}$, supporting its use as a proxy for visual reliance.

We then set $v_{\mathrm{deg}}$ to a black image $v_{\mathrm{blind}}$ and vary the proportion of similarity-retrieved and randomly selected demonstrations in the HC-ICL context.

\begin{figure}[t]
    \centering
    \includegraphics[width=\linewidth]{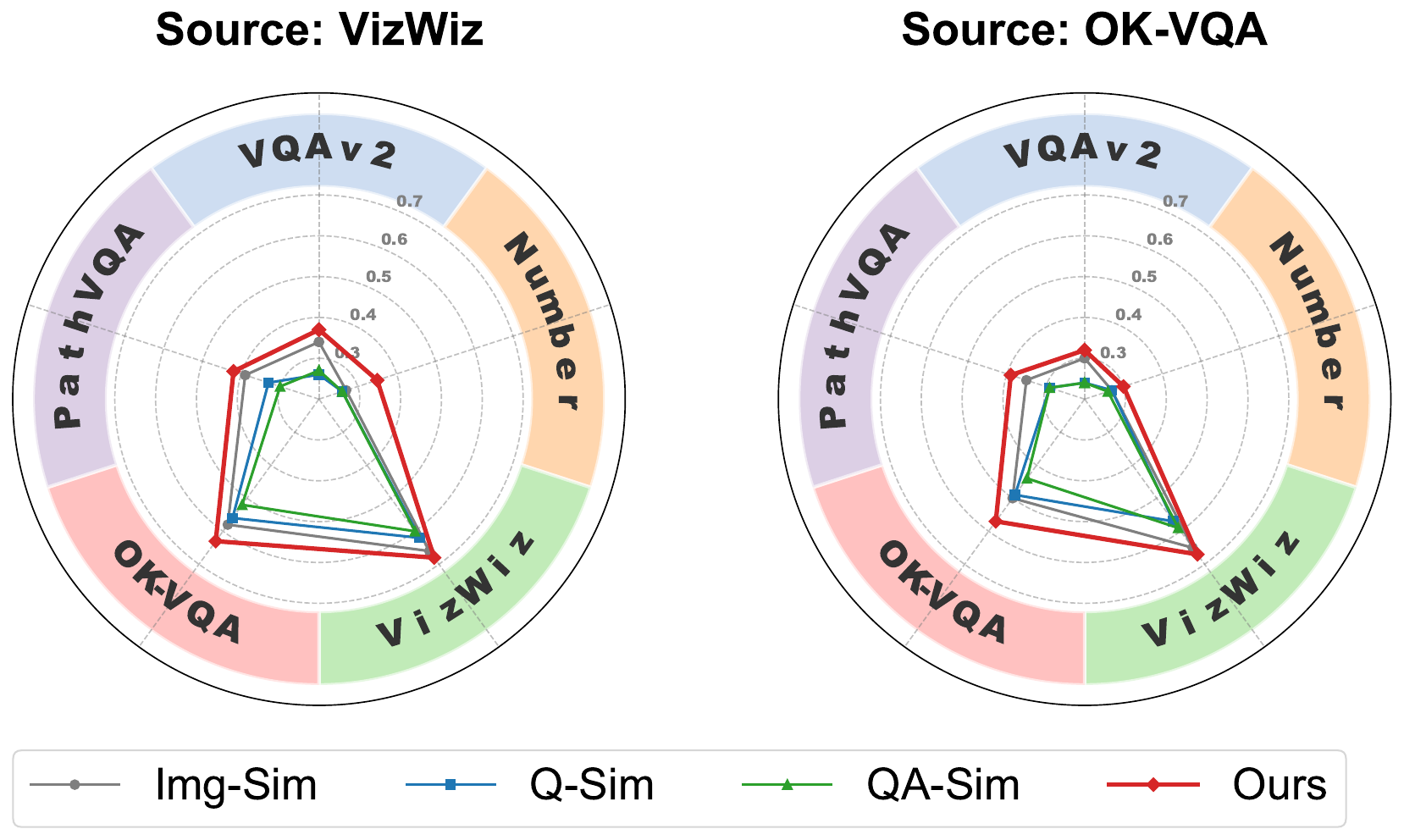}
    \caption{Comparison of visual reliance ($D_{\mathrm{KL}}$) across different selection strategies under the HC-ICL setting.}
    \label{fig:kl_compare}
\end{figure}

Figure~\ref{fig:kl_all} shows that increasing the proportion of
similarity-retrieved demonstrations consistently reduces
$D_{\mathrm{KL}}$ across source--target pairs. In-domain accuracy remains
stable because aligned textual patterns compensate for reduced visual
reliance, whereas under shifts such as Number $\rightarrow$ VizWiz, the
same effect causes substantial degradation. Thus, similarity-induced
semantic redundancy weakens visual reliance and is particularly harmful
under domain shifts.

Figure~\ref{fig:kl_compare} further compares visual reliance across different context-selection strategies. Our method consistently maintains higher $D_{\mathrm{KL}}$ values than the similarity-based baselines across source domains, indicating that it better preserves reliance on target-image evidence in cross-modal reasoning.

\begin{figure}[t]
    \centering
    \includegraphics[width=\linewidth]{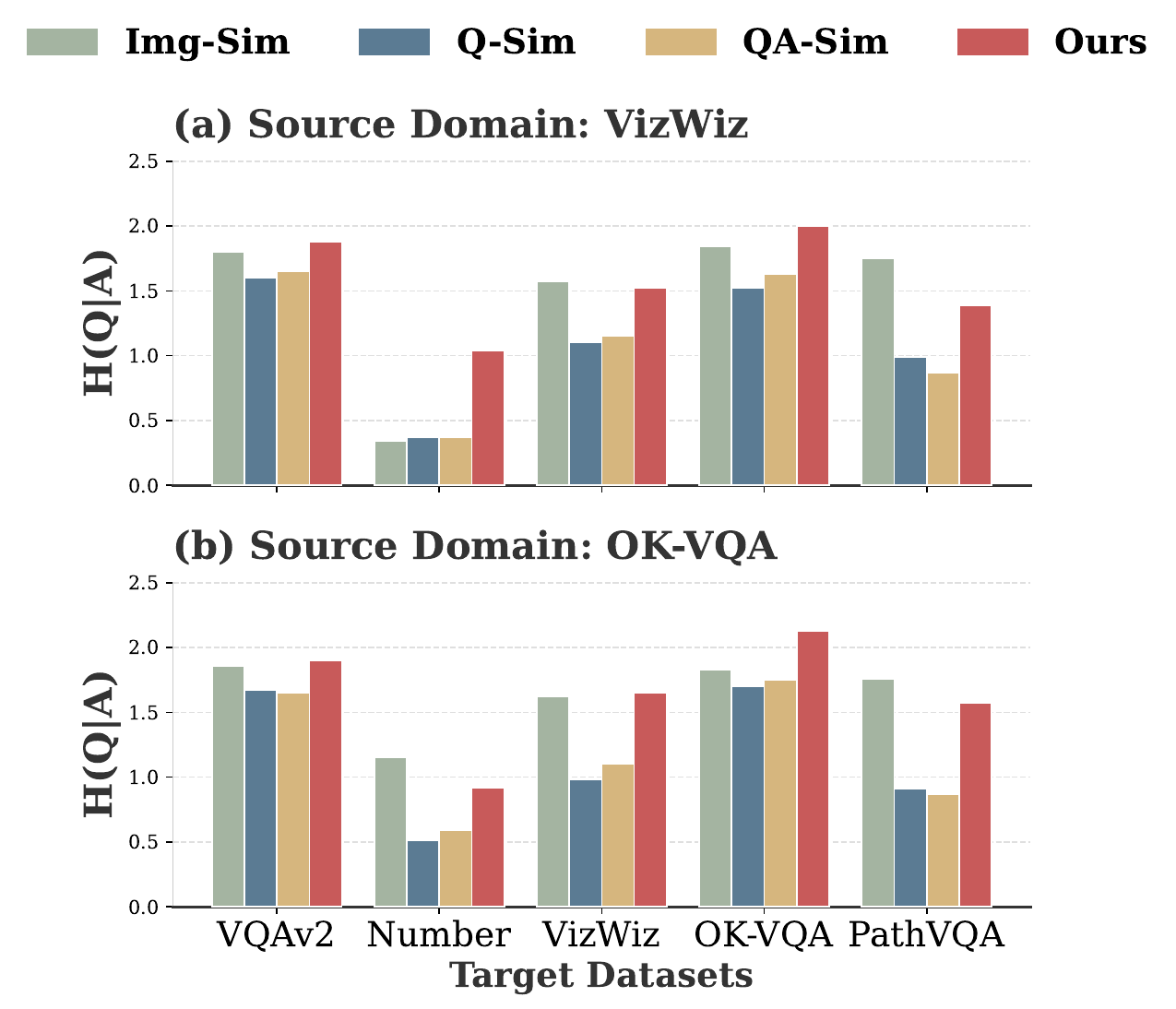}
    \caption{Comparison of conditional entropy $H(Q|A)$ across different selection strategies and target datasets under the HC-ICL setting.}
    \label{fig:entropy_analysis}
\end{figure}

\noindent\textbf{Structural Diversity Mitigates Spurious Shortcuts.}
We quantify demonstration diversity using conditional entropy $H(Q\mid A)$, where questions are grouped into intent clusters according to functional prefixes such as ``What is'' and ``How many.'' Unlike unconditional entropy $H(Q)$, this metric captures structural variation while preserving answer-conditioned patterns.

As shown in Figure~\ref{fig:entropy_analysis}, QA-Sim produces low $H(Q\mid A)$ because it retrieves repetitive question structures, whereas Img-Sim provides greater diversity and our method achieves the highest entropy across cross-domain settings. These results suggest that answer-conditioned structural diversity suppresses textual shortcuts and promotes greater reliance on target-image evidence.
\section{Conclusion}
We formulate Hybrid-Context ICL, a practical setting where demonstrations combine labeled OOD samples with pseudo-labeled target-domain samples, exposing a fundamental trade-off between label reliability and distributional alignment. Our analysis shows that similarity-based retrieval reinforces repetitive textual patterns, reduces reliance on target-image evidence, and becomes particularly unreliable under domain shifts. To address these limitations, we propose Progressive Arbitration of Contextual Evidence, which suppresses target-image-independent shortcuts and generic source-context bias while progressively incorporating consistent and reliable target-domain evidence through a dynamic pool. The framework thus gradually shifts the demonstration context from source-biased but label-reliable evidence toward target-compatible support without updating model parameters. Experiments across diverse VQA datasets and source--target configurations show consistent gains over existing retrieval and adaptation baselines, while analyses confirm that the method better preserves visual reliance and structural diversity under imperfect context conditions.

\bibliography{aaai2027}

@article{brown2020language,
  title={Language models are few-shot learners},
  author={Brown, Tom and Mann, Benjamin and Ryder, Nick and Subbiah, Melanie and Kaplan, Jared D and Dhariwal, Prafulla and Neelakantan, Arvind and Shyam, Pranav and Sastry, Girish and Askell, Amanda and others},
  journal={Advances in neural information processing systems},
  volume={33},
  pages={1877--1901},
  year={2020}
}

@article{wei2022chain,
  title={Chain-of-thought prompting elicits reasoning in large language models},
  author={Wei, Jason and Wang, Xuezhi and Schuurmans, Dale and Bosma, Maarten and Xia, Fei and Chi, Ed and Le, Quoc V and Zhou, Denny and others},
  journal={Advances in neural information processing systems},
  volume={35},
  pages={24824--24837},
  year={2022}
}

@article{alayrac2022flamingo,
  title={Flamingo: a visual language model for few-shot learning},
  author={Alayrac, Jean-Baptiste and Donahue, Jeff and Luc, Pauline and Miech, Antoine and Barr, Iain and Hasson, Yana and Lenc, Karel and Mensch, Arthur and Millican, Katherine and Reynolds, Malcolm and others},
  journal={Advances in neural information processing systems},
  volume={35},
  pages={23716--23736},
  year={2022}
}

@article{awadalla2023openflamingo,
  title={OpenFlamingo: An open-source framework for training large autoregressive vision-language models},
  author={Awadalla, Anas and Gao, Irena and Gardner, Josh and Hessel, Jack and Hanafy, Yusuf and Zhu, Wanrong and Marathe, Kalyani and Bitton, Yonatan and Gadre, Samir and Sagawa, Shiori and others},
  journal={arXiv preprint arXiv:2308.01390},
  year={2023}
}

@inproceedings{doveh2024towards,
  title={Towards multimodal in-context learning for vision and language models},
  author={Doveh, Sivan and Perek, Shaked and Mirza, M Jehanzeb and Lin, Wei and Alfassy, Amit and Arbelle, Assaf and Ullman, Shimon and Karlinsky, Leonid},
  booktitle={European Conference on Computer Vision},
  pages={250--267},
  year={2024},
  organization={Springer}
}

@inproceedings{zhao2021calibrate,
  title={Calibrate before use: Improving few-shot performance of language models},
  author={Zhao, Zihao and Wallace, Eric and Feng, Shi and Klein, Dan and Singh, Sameer},
  booktitle={International conference on machine learning},
  pages={12697--12706},
  year={2021},
  organization={Pmlr}
}

@inproceedings{liu2022makes,
  title={What makes good in-context examples for GPT-3?},
  author={Liu, Jiachang and Shen, Dinghan and Zhang, Yizhe and Dolan, William B and Carin, Lawrence and Chen, Weizhu},
  booktitle={Proceedings of Deep Learning Inside Out (DeeLIO 2022): The 3rd workshop on knowledge extraction and integration for deep learning architectures},
  pages={100--114},
  year={2022}
}

@inproceedings{li2024configure,
  title={How to configure good in-context sequence for visual question answering},
  author={Li, Li and Peng, Jiawei and Chen, Huiyi and Gao, Chongyang and Yang, Xu},
  booktitle={Proceedings of the IEEE/CVF Conference on Computer Vision and Pattern Recognition},
  pages={26710--26720},
  year={2024}
}

@inproceedings{zhou2024explore,
  title={Explore spurious correlations at the concept level in language models for text classification},
  author={Zhou, Yuhang and Xu, Paiheng and Liu, Xiaoyu and An, Bang and Ai, Wei and Huang, Furong},
  booktitle={Proceedings of the 62nd Annual Meeting of the Association for Computational Linguistics (Volume 1: Long Papers)},
  pages={478--492},
  year={2024}
}

@inproceedings{baldassini2024makes,
  title={What makes multimodal in-context learning work?},
  author={Baldassini, Folco Bertini and Shukor, Mustafa and Cord, Matthieu and Soulier, Laure and Piwowarski, Benjamin},
  booktitle={Proceedings of the IEEE/CVF Conference on Computer Vision and Pattern Recognition},
  pages={1539--1550},
  year={2024}
}

@inproceedings{zhang2022active,
  title={Active example selection for in-context learning},
  author={Zhang, Yiming and Feng, Shi and Tan, Chenhao},
  booktitle={Proceedings of the 2022 Conference on Empirical Methods in Natural Language Processing},
  pages={9134--9148},
  year={2022}
}

@inproceedings{mo2024c,
  title={C-ICL: contrastive in-context learning for information extraction},
  author={Mo, Ying and Liu, Jiahao and Yang, Jian and Wang, Qifan and Zhang, Shun and Wang, Jingang and Li, Zhoujun},
  booktitle={Findings of the Association for Computational Linguistics: EMNLP 2024},
  pages={10099--10114},
  year={2024}
}

@article{nanda2025insquad,
  title={InSQuAD: In-Context Learning for Efficient Retrieval via Submodular Mutual Information to Enforce Quality and Diversity},
  author={Nanda, Souradeep and Majee, Anay and Iyer, Rishabh},
  journal={arXiv preprint arXiv:2508.21003},
  year={2025}
}

@inproceedings{li2023blip,
  title={Blip-2: Bootstrapping language-image pre-training with frozen image encoders and large language models},
  author={Li, Junnan and Li, Dongxu and Savarese, Silvio and Hoi, Steven},
  booktitle={International conference on machine learning},
  pages={19730--19742},
  year={2023},
  organization={PMLR}
}

@article{liu2023visual,
  title={Visual instruction tuning},
  author={Liu, Haotian and Li, Chunyuan and Wu, Qingyang and Lee, Yong Jae},
  journal={Advances in neural information processing systems},
  volume={36},
  pages={34892--34916},
  year={2023}
}

@article{wang2024qwen2,
  title={Qwen2-vl: Enhancing vision-language model's perception of the world at any resolution},
  author={Wang, Peng and Bai, Shuai and Tan, Sinan and Wang, Shijie and Fan, Zhihao and Bai, Jinze and Chen, Keqin and Liu, Xuejing and Wang, Jialin and Ge, Wenbin and others},
  journal={arXiv preprint arXiv:2409.12191},
  year={2024}
}

@article{gadetsky2025large,
  title={Large (vision) language models are unsupervised in-context learners},
  author={Gadetsky, Artyom and Atanov, Andrei and Jiang, Yulun and Gao, Zhitong and Mighan, Ghazal Hosseini and Zamir, Amir and Brbic, Maria},
  journal={arXiv preprint arXiv:2504.02349},
  year={2025}
}

@misc{mosaicml2023introducing,
  title={Introducing mpt-7b: A new standard for open-source, commercially usable llms},
  author={MosaicML NLP Team and others},
  year={2023},
  publisher={Accessed}
}

@inproceedings{radford2021learning,
  title={Learning transferable visual models from natural language supervision},
  author={Radford, Alec and Kim, Jong Wook and Hallacy, Chris and Ramesh, Aditya and Goh, Gabriel and Agarwal, Sandhini and Sastry, Girish and Askell, Amanda and Mishkin, Pamela and Clark, Jack and others},
  booktitle={International conference on machine learning},
  pages={8748--8763},
  year={2021},
  organization={PmLR}
}

@InProceedings{Goyal_2017_CVPR,
author = {Goyal, Yash and Khot, Tejas and Summers-Stay, Douglas and Batra, Dhruv and Parikh, Devi},
title = {Making the v in VQA Matter: Elevating the Role of Image Understanding in Visual Question Answering},
booktitle = {Proceedings of the IEEE Conference on Computer Vision and Pattern Recognition (CVPR)},
month = {July},
year = {2017}
}

@inproceedings{marino2019ok,
  title={Ok-vqa: A visual question answering benchmark requiring external knowledge},
  author={Marino, Kenneth and Rastegari, Mohammad and Farhadi, Ali and Mottaghi, Roozbeh},
  booktitle={Proceedings of the IEEE/cvf conference on computer vision and pattern recognition},
  pages={3195--3204},
  year={2019}
}

@InProceedings{Gurari_2018_CVPR,
author = {Gurari, Danna and Li, Qing and Stangl, Abigale J. and Guo, Anhong and Lin, Chi and Grauman, Kristen and Luo, Jiebo and Bigham, Jeffrey P.},
title = {VizWiz Grand Challenge: Answering Visual Questions From Blind People},
booktitle = {Proceedings of the IEEE Conference on Computer Vision and Pattern Recognition (CVPR)},
month = {June},
year = {2018}
}

@article{he2020pathvqa,
  title={PathVQA: 30000+ questions for medical visual question answering},
  author={He, Xuehai and Zhang, Yichen and Mou, Luntian and Xing, Eric and Xie, Pengtao},
  journal={arXiv preprint arXiv:2003.10286},
  year={2020}
}

@article{ren2015exploring,
  title={Exploring models and data for image question answering},
  author={Ren, Mengye and Kiros, Ryan and Zemel, Richard},
  journal={Advances in neural information processing systems},
  volume={28},
  year={2015}
}

@article{olsson2022context,
  title={In-context learning and induction heads},
  author={Olsson, Catherine and Elhage, Nelson and Nanda, Neel and Joseph, Nicholas and DasSarma, Nova and Henighan, Tom and Mann, Ben and Askell, Amanda and Bai, Yuntao and Chen, Anna and others},
  journal={arXiv preprint arXiv:2209.11895},
  year={2022}
}

@article{ou2026mllms,
  title={Do MLLMs Really See It: Reinforcing Visual Attention in Multimodal LLMs},
  author={Ou, Siqu and Wan, Tianrui and Zhao, Zhiyuan and Gao, Junyu and Li, Xuelong},
  journal={arXiv preprint arXiv:2602.08241},
  year={2026}
}

@inproceedings{schwettmann2023multimodal,
  title={Multimodal neurons in pretrained text-only transformers},
  author={Schwettmann, Sarah and Chowdhury, Neil and Klein, Samuel and Bau, David and Torralba, Antonio},
  booktitle={Proceedings of the IEEE/CVF International Conference on Computer Vision},
  pages={2862--2867},
  year={2023}
}

@article{goddard2025can,
  title={When can in-context learning generalize out of task distribution?},
  author={Goddard, Chase and Smith, Lindsay M and Ngampruetikorn, Vudtiwat and Schwab, David J},
  journal={arXiv preprint arXiv:2506.05574},
  year={2025}
}

@article{kwon2025out,
  title={Out-of-distribution generalization of in-context learning: A low-dimensional subspace perspective},
  author={Kwon, Soo Min and Xu, Alec S and Yaras, Can and Balzano, Laura and Qu, Qing},
  journal={arXiv preprint arXiv:2505.14808},
  year={2025}
}

@article{song2025out,
  title={Out-of-distribution generalization via composition: a lens through induction heads in transformers},
  author={Song, Jiajun and Xu, Zhuoyan and Zhong, Yiqiao},
  journal={Proceedings of the National Academy of Sciences},
  volume={122},
  number={6},
  pages={e2417182122},
  year={2025},
  publisher={National Academy of Sciences}
}

@article{peng2024live,
  title={Live: Learnable in-context vector for visual question answering},
  author={Peng, Yingzhe and Hao, Chenduo and Hu, Xinting and Peng, Jiawei and Geng, Xin and Yang, Xu},
  journal={Advances in Neural Information Processing Systems},
  volume={37},
  pages={9773--9800},
  year={2024}
}

@article{dang2024sadl,
  title={SADL: An effective in-context learning method for compositional visual qa},
  author={Dang, Long Hoang and Le, Thao Minh and Le, Vuong and Phuong, Tu Minh and Tran, Truyen},
  journal={arXiv preprint arXiv:2407.01983},
  year={2024}
}

@article{liu2022deep,
  title={Deep unsupervised domain adaptation: A review of recent advances and perspectives},
  author={Liu, Xiaofeng and Yoo, Chaehwa and Xing, Fangxu and Oh, Hyejin and Fakhri, Georges El and Kang, Je-Won and Woo, Jonghye},
  journal={arXiv preprint arXiv:2208.07422},
  year={2022}
}

@inproceedings{chen2022contrastive,
  title={Contrastive test-time adaptation},
  author={Chen, Dian and Wang, Dequan and Darrell, Trevor and Ebrahimi, Sayna},
  booktitle={Proceedings of the IEEE/CVF conference on computer vision and pattern recognition},
  pages={295--305},
  year={2022}
}

@InProceedings{Antol_2015_ICCV,
author = {Antol, Stanislaw and Agrawal, Aishwarya and Lu, Jiasen and Mitchell, Margaret and Batra, Dhruv and Zitnick, C. Lawrence and Parikh, Devi},
title = {VQA: Visual Question Answering},
booktitle = {Proceedings of the IEEE International Conference on Computer Vision (ICCV)},
month = {December},
year = {2015}
}

@inproceedings{gao2021making,
  title={Making pre-trained language models better few-shot learners},
  author={Gao, Tianyu and Fisch, Adam and Chen, Danqi},
  booktitle={Proceedings of the 59th Annual Meeting of the Association for Computational Linguistics and the 11th International Joint Conference on Natural Language Processing (Volume 1: Long Papers)},
  pages={3816--3830},
  year={2021}
}

@article{ganin2016domain,
  title={Domain-adversarial training of neural networks},
  author={Ganin, Yaroslav and Ustinova, Evgeniya and Ajakan, Hana and Germain, Pascal and Larochelle, Hugo and Laviolette, Fran{\c{c}}ois and March, Mario and Lempitsky, Victor},
  journal={Journal of machine learning research},
  volume={17},
  number={59},
  pages={1--35},
  year={2016}
}

@article{garg2022can,
  title={What can transformers learn in-context? a case study of simple function classes},
  author={Garg, Shivam and Tsipras, Dimitris and Liang, Percy S and Valiant, Gregory},
  journal={Advances in neural information processing systems},
  volume={35},
  pages={30583--30598},
  year={2022}
}

@article{wang2024can,
  title={Can In-context Learning Really Generalize to Out-of-distribution Tasks?},
  author={Wang, Qixun and Wang, Yifei and Wang, Yisen and Ying, Xianghua},
  journal={arXiv preprint arXiv:2410.09695},
  year={2024}
}

@article{laskin2022context,
  title={In-context reinforcement learning with algorithm distillation},
  author={Laskin, Michael and Wang, Luyu and Oh, Junhyuk and Parisotto, Emilio and Spencer, Stephen and Steigerwald, Richie and Strouse, DJ and Hansen, Steven and Filos, Angelos and Brooks, Ethan and others},
  journal={arXiv preprint arXiv:2210.14215},
  year={2022}
}

@inproceedings{gao2024customizing,
  title={Customizing language model responses with contrastive in-context learning},
  author={Gao, Xiang and Das, Kamalika},
  booktitle={Proceedings of the aaai conference on artificial intelligence},
  volume={38},
  number={16},
  pages={18039--18046},
  year={2024}
}

@inproceedings{wu2023self,
  title={Self-adaptive in-context learning: An information compression perspective for in-context example selection and ordering},
  author={Wu, Zhiyong and Wang, Yaoxiang and Ye, Jiacheng and Kong, Lingpeng},
  booktitle={Proceedings of the 61st Annual Meeting of the Association for Computational Linguistics (Volume 1: Long Papers)},
  pages={1423--1436},
  year={2023}
}

@inproceedings{chen2025can,
  title={Can Multimodal Large Language Models Truly Perform Multimodal In-Context Learning?},
  author={Chen, Shuo and Han, Zhen and He, Bailan and Liu, Jianzhe and Buckley, Mark and Qin, Yao and Torr, Philip and Tresp, Volker and Gu, Jindong},
  booktitle={2025 IEEE/CVF Winter Conference on Applications of Computer Vision (WACV)},
  pages={6000--6010},
  year={2025},
  organization={IEEE}
}

@article{chen2025true,
  title={True multimodal in-context learning needs attention to the visual context},
  author={Chen, Shuo and Liu, Jianzhe and Han, Zhen and Xia, Yan and Cremers, Daniel and Torr, Philip and Tresp, Volker and Gu, Jindong},
  journal={arXiv preprint arXiv:2507.15807},
  year={2025}
}

@inproceedings{zhou2025bayesian,
  title={Bayesian test-time adaptation for vision-language models},
  author={Zhou, Lihua and Ye, Mao and Li, Shuaifeng and Li, Nianxin and Zhu, Xiatian and Deng, Lei and Liu, Hongbin and Lei, Zhen},
  booktitle={Proceedings of the Computer Vision and Pattern Recognition Conference},
  pages={29999--30009},
  year={2025}
}

@inproceedings{fan2025test,
  title={Test-Time Retrieval-Augmented Adaptation for Vision-Language Models},
  author={Fan, Xinqi and Chen, Xueli and Yang, Luoxiao and Yap, Chuin Hong and Qureshi, Rizwan and Dou, Qi and Yap, Moi Hoon and Shah, Mubarak},
  booktitle={Proceedings of the IEEE/CVF International Conference on Computer Vision},
  pages={8810--8819},
  year={2025}
}

@article{xie2021explanation,
  title={An explanation of in-context learning as implicit bayesian inference},
  author={Xie, Sang Michael and Raghunathan, Aditi and Liang, Percy and Ma, Tengyu},
  journal={arXiv preprint arXiv:2111.02080},
  year={2021}
}

@article{zhang2023and,
  title={What and how does in-context learning learn? bayesian model averaging, parameterization, and generalization},
  author={Zhang, Yufeng and Zhang, Fengzhuo and Yang, Zhuoran and Wang, Zhaoran},
  journal={arXiv preprint arXiv:2305.19420},
  year={2023}
}

@inproceedings{panwar2024context,
  title={In-context learning through the bayesian prism},
  author={Panwar, Madhur and Ahuja, Kabir and Goyal, Navin},
  booktitle={International Conference on Learning Representations},
  volume={2024},
  pages={49789--49843},
  year={2024}
}

@article{grattafiori2024llama,
  title={The llama 3 herd of models},
  author={Grattafiori, Aaron and Dubey, Abhimanyu and Jauhri, Abhinav and Pandey, Abhinav and Kadian, Abhishek and Al-Dahle, Ahmad and Letman, Aiesha and Mathur, Akhil and Schelten, Alan and Vaughan, Alex and others},
  journal={arXiv preprint arXiv:2407.21783},
  year={2024}
}

@inproceedings{zhai2023sigmoid,
  title={Sigmoid loss for language image pre-training},
  author={Zhai, Xiaohua and Mustafa, Basil and Kolesnikov, Alexander and Beyer, Lucas},
  booktitle={Proceedings of the IEEE/CVF international conference on computer vision},
  pages={11975--11986},
  year={2023}
}

\clearpage
\appendix
\setcounter{figure}{0}
\setcounter{table}{0}
\setcounter{equation}{0}
\renewcommand{\thefigure}{S\arabic{figure}}
\renewcommand{\thetable}{S\arabic{table}}
\renewcommand{\theequation}{S\arabic{equation}}

\twocolumn[{
\begin{center}
{\Large\bfseries Supplementary Materials\par}
\vspace{0.5em}

\end{center}
\vspace{1em}
}]

\section{Method}

This section provides an extended technical description of the
Progressive Arbitration of Contextual Evidence framework. We formalize the
coarse retrieval stage, progressive context construction, and
adaptive admission mechanism described in the main paper.

\noindent\textbf{Coarse Retrieval and Candidate Initialization.}
Our Progressive Arbitration of Contextual Evidence framework follows a retrieve-then-rerank
pipeline. Before applying Dual-Prior Contrastive Scoring (DPCS),
we retrieve an initial candidate subset from the heterogeneous
hybrid pool
$\mathcal{S}_h=\mathcal{S}_l\cup\widehat{\mathcal{S}}_u$,
where $\mathcal{S}_l$ is the labeled source-domain pool and
$\widehat{\mathcal{S}}_u$ is the target-domain pool equipped with
model-generated pseudo-labels.

For a target query, we define
$s_t=(v_t,q_t,\hat{a}_t^{(0)})$, where
$\hat{a}_t^{(0)}$ denotes the initial pseudo-label generated by
the base model under a labeled source-domain ICL context. A support instance is written as
$s_i=(v_i,q_i,a_i)\in\mathcal{S}_h$, where $a_i$ is either a
source-domain ground-truth label or a target-domain pseudo-label.
Let $\phi(\cdot)$ map the modalities used by a retrieval strategy,
such as the image, question, or question-answer pair, into a
$d$-dimensional representation. We denote the resulting features
by $h_t=\phi(s_t)$ and $h_i=\phi(s_i)$. When answer information
is not used, the corresponding component is simply ignored by
$\phi(\cdot)$.

The initial candidate pool $\mathcal{E}$ contains the
top-$K_{\mathrm{recall}}$ samples with the highest cosine
similarity to the target query:

\begin{equation}
\mathcal{E}
=
\operatorname*{arg\,max}_{
\substack{
\mathcal{E}'\subset\mathcal{S}_h\\
|\mathcal{E}'|=K_{\mathrm{recall}}
}
}
\sum_{s_i\in\mathcal{E}'}
\frac{h_t\cdot h_i}
{\lVert h_t\rVert\,\lVert h_i\rVert}.
\tag{a}
\end{equation}

This coarse retrieval stage restricts contrastive reranking to
candidates with a baseline level of surface relevance, allowing
DPCS to focus on separating useful predictive evidence from
spurious shortcuts.

\noindent\textbf{Formalization of Progressive Contextual Evolution.}
Let $K$ denote the required number of context demonstrations, and
let $\mathcal{E}_{\mathrm{valid}}$ denote the candidates retained
by DPCS in descending score order. Because the number of retained
candidates may exceed the required context length, we keep only
the highest-ranked candidates, denoted by
$\mathcal{E}_K=
\operatorname{Top}_{\min(K,|\mathcal{E}_{\mathrm{valid}}|)}
(\mathcal{E}_{\mathrm{valid}})$.
The remaining contextual gap is
$\Delta_K=K-|\mathcal{E}_K|$.

At inference step $t$, the system maintains a dynamic target-domain
pool $\mathcal{S}_{\mathrm{dyn}}^{(t)}$. To avoid duplicate
demonstrations, we exclude samples already selected in
$\mathcal{E}_K$ and define
$\mathcal{D}_t=
\mathcal{S}_{\mathrm{dyn}}^{(t)}\setminus\mathcal{E}_K$
and
$\mathcal{L}_t=
\mathcal{S}_l\setminus\mathcal{E}_K$.
Let $n_t=|\mathcal{D}_t|$, and abbreviate
$\operatorname{Sample}$ as $\operatorname{Samp}$ in the following
formulation. The supplementary context is constructed as

\begin{equation}
C_{\mathrm{supp}}
\sim
\begin{cases}
[\,], & \Delta_K=0, \\[1mm]

\operatorname{Samp}(\Delta_K,\mathcal{L}_t),
& n_t=0, \\[1mm]

\begin{aligned}
&\operatorname{Samp}(n_t,\mathcal{D}_t)\,\Vert\,{}\\[-1mm]
&\operatorname{Samp}(\Delta_K-n_t,\mathcal{L}_t)
\end{aligned}
& 0<n_t<\Delta_K, \\[3mm]

\operatorname{Samp}(\Delta_K,\mathcal{D}_t),
& n_t\geq\Delta_K .
\end{cases}
\tag{b}
\end{equation}

Here, $\operatorname{Samp}(m,\mathcal{S})$ returns an ordered
sequence of $m$ distinct samples drawn without replacement from
$\mathcal{S}$, and $\Vert$ denotes sequence concatenation. The
final context is constructed as
$C_t=\mathcal{E}_K\Vert C_{\mathrm{supp}}$.
This construction prioritizes the high-gain candidates retained
by DPCS and fills the remaining positions using target-aligned
samples from the dynamic pool. When sufficient reliable
target-domain samples have not yet been accumulated, labeled
source-domain samples provide a conservative fallback during
cold start.

\noindent\textbf{Dual Adaptive Gating.}
For the current target query $(v_t,q_t)$, let
$\hat{a}_t^{(0)}$ denote its initial pseudo-label generated under
the labeled source-domain ICL context, and let
$\hat{a}_t^{(C_t)}=\mathcal{M}(C_t,v_t,q_t)$ denote the prediction
decoded under the constructed hybrid context. We use $\gamma_t$ to denote
the corresponding generation confidence, computed as the maximum
softmax probability at the first decoding step.

A target-domain sample is admitted into the dynamic pool only
when the initial and contextual predictions are consistent and
the contextual prediction is sufficiently confident:

\begin{equation}
\mathbb{I}_{\mathrm{admit}}
=
\mathbb{I}\!\left(
\hat{a}_t^{(C_t)}=\hat{a}_t^{(0)}
\right)
\land
\mathbb{I}\!\left(
\gamma_t\geq\tau_t
\right).
\tag{c}
\end{equation}

The consistency condition provides a form of self-verification,
while the confidence condition suppresses uncertain pseudo-labels.
If $\mathbb{I}_{\mathrm{admit}}=1$, the sample
$(v_t,q_t,\hat{a}_t^{(C_t)})$ is added to
$\mathcal{S}_{\mathrm{dyn}}^{(t)}$; otherwise, the dynamic pool
remains unchanged.

The threshold used at the next inference step is updated as
$\tau_{t+1}=
\max(\tau_{\min},\mu_t^{(W)}-\sigma_t^{(W)})$, where
$\mu_t^{(W)}$ and $\sigma_t^{(W)}$ are computed from the
generation confidences of the most recent $W$ admitted samples.
Before any sample is admitted, we initialize
$\tau_0=\tau_{\min}$. When fewer than $W$ admitted samples are
available, the statistics are computed over all currently
admitted samples.

\begin{table}[t]
\centering
\footnotesize

\begin{tabular*}{\columnwidth}{@{\extracolsep{\fill}} lcccc @{}}
\toprule
\multicolumn{5}{c}{\textbf{Source Domain: VQAv2}} \\
\midrule
\textbf{Target} & \textbf{1-shot} & \textbf{2-shot} & \textbf{4-shot} & \textbf{8-shot} \\
\midrule
VQAv2   & 36.23 & 44.99 & 49.94 & \textbf{52.62} \\
Number  & 36.02 & 36.63 & 39.79 & \textbf{39.97} \\
VizWiz  & 16.50 & 17.42 & 19.85 & \textbf{21.16} \\
OK-VQA  & 20.20 & 25.35 & 30.26 & \textbf{34.91} \\
PathVQA & 15.36 & 18.02 & 23.89 & \textbf{27.74} \\
Avg.    & 24.86 & 28.48 & 32.75 & \textbf{35.28} \\
\midrule
\multicolumn{5}{c}{\textbf{Source Domain: Number}} \\
\midrule
\textbf{Target} & \textbf{1-shot} & \textbf{2-shot} & \textbf{4-shot} & \textbf{8-shot} \\
\midrule
VQAv2   & 22.24 & 34.40 & 45.61 & \textbf{55.79} \\
Number  & 46.43 & \textbf{48.28} & 46.05 & 44.15 \\
VizWiz  & 11.10 & 13.14 & 14.48 & \textbf{17.54} \\
OK-VQA  & 17.06 & 23.02 & 28.20 & \textbf{33.08} \\
PathVQA & 10.05 & 20.26 & 25.92 & \textbf{30.67} \\
Avg.    & 21.38 & 27.82 & 32.05 & \textbf{36.25} \\
\bottomrule
\end{tabular*}
\caption{Cross-domain evaluation of PACE under varying few-shot sizes (Accuracy, \%). The table reports performance across the all-to-all dataset matrix for $K \in \{1, 2, 4, 8\}$. Results indicate that the average accuracy generally improves as the number of shots increases. Best results within each source domain group are in bold.}
\label{supp:tab:shot}
\end{table}

\begin{table}[t]
\ContinuedFloat
\centering
\footnotesize

\begin{tabular*}{\columnwidth}{@{\extracolsep{\fill}} lcccc @{}}
\toprule
\multicolumn{5}{c}{\textbf{Source Domain: VizWiz}} \\
\midrule
\textbf{Target} & \textbf{1-shot} & \textbf{2-shot} & \textbf{4-shot} & \textbf{8-shot} \\
\midrule
VQAv2   & 17.12 & 29.55 & 42.20 & \textbf{48.45} \\
Number  & 38.77 & 42.18 & \textbf{43.00} & 42.99 \\
VizWiz  & 45.69 & 47.26 & 49.18 & \textbf{50.50} \\
OK-VQA  & 11.72 & 18.35 & 22.76 & \textbf{28.81} \\
PathVQA & 15.83 & 17.58 & 19.80 & \textbf{22.12} \\
Avg.    & 25.83 & 30.98 & 35.39 & \textbf{38.57} \\
\midrule
\multicolumn{5}{c}{\textbf{Source Domain: OK-VQA}} \\
\midrule
\textbf{Target} & \textbf{1-shot} & \textbf{2-shot} & \textbf{4-shot} & \textbf{8-shot} \\
\midrule
VQAv2   & 19.08 & 33.46 & 42.25 & \textbf{45.04} \\
Number  & 34.42 & 39.20 & \textbf{43.58} & 39.93 \\
VizWiz  & 14.16 & 17.02 & 18.38 & \textbf{21.13} \\
OK-VQA  & 32.61 & 35.28 & 36.60 & \textbf{37.01} \\
PathVQA & 7.03  & 14.00 & 21.57 & \textbf{24.13} \\
Avg.    & 21.46 & 27.79 & 32.48 & \textbf{33.45} \\
\midrule
\multicolumn{5}{c}{\textbf{Source Domain: PathVQA}} \\
\midrule
\textbf{Target} & \textbf{1-shot} & \textbf{2-shot} & \textbf{4-shot} & \textbf{8-shot} \\
\midrule
VQAv2   & 28.56 & 32.87 & \textbf{36.55} & 35.99 \\
Number  & 35.84 & 37.08 & 39.00 & \textbf{39.77} \\
VizWiz  & 6.37  & 10.81 & 12.83 & \textbf{14.64} \\
OK-VQA  & 13.83 & 19.31 & 24.61 & \textbf{26.94} \\
PathVQA & 36.72 & 28.55 & 38.90 & \textbf{41.92} \\
Avg.    & 24.26 & 25.72 & 30.38 & \textbf{31.85} \\
\bottomrule
\end{tabular*}
\caption[]{Cross-domain evaluation under varying few-shot sizes (continued).}
\end{table}


\begin{table*}[t]
\centering

\small
\setlength{\tabcolsep}{3.5pt}

\begin{tabular*}{\textwidth}{
@{\extracolsep{\fill}}
llcccccc
@{}
}
\toprule
\textbf{Method}
& \textbf{Source}
& \textbf{VQAv2}
& \textbf{Number}
& \textbf{VizWiz}
& \textbf{OK-VQA}
& \textbf{PathVQA}
& \textbf{Avg.} \\
\midrule

\multirow{5}{*}{\textbf{Base}}
& VQAv2
& 48.29 & 35.60 & 19.29 & 27.31 & 20.79 & 30.26 \\

& Number
& 42.75 & 47.11 & 16.86 & 24.17 & 25.61 & 31.30 \\

& VizWiz
& 30.63 & 33.45 & 35.56 & 19.63 & 9.40 & 25.73 \\

& OK-VQA
& 35.92 & 35.86 & \textbf{20.19} & 36.67 & 14.19 & 28.57 \\

& PathVQA
& 27.33 & 34.35 & 10.90 & 18.48 & 38.47 & 25.91 \\
\midrule

\multirow{5}{*}{\textbf{TT-RAA}}
& VQAv2
& 48.06 & 25.31 & 16.91 & 30.11 & 15.04 & 27.09 \\

& Number
& 28.20 & 36.88 & 12.75 & 18.54 & 24.37 & 24.15 \\

& VizWiz
& 20.81 & 19.19 & 38.88 & 18.45 & 5.71 & 20.61 \\

& OK-VQA
& 30.16 & 11.74 & 15.23 & 35.16 & 9.68 & 20.39 \\

& PathVQA
& 14.34 & 5.26 & 7.27 & 15.23 & 34.04 & 15.23 \\
\midrule

\multirow{5}{*}{\textbf{U-ICL}}
& VQAv2
& 43.97 & 37.34 & 17.83 & 29.80 & 19.03 & 29.59 \\

& Number
& 40.70 & 38.13 & \textbf{19.04} & 20.88 & 24.27 & 28.60 \\

& VizWiz
& 40.19 & 35.32 & 40.93 & 24.63 & 11.59 & 30.53 \\

& OK-VQA
& 39.03 & 34.26 & 13.84 & 35.33 & 12.26 & 26.94 \\

& PathVQA
& 29.31 & 38.00 & 9.02 & 22.18 & 35.60 & 26.82 \\
\midrule

\multirow{5}{*}{\textbf{BCA}}
& VQAv2
& 48.36 & 38.19 & 19.22 & 28.59 & 19.29 & 30.73 \\

& Number
& 43.03 & 44.12 & 16.43 & 26.98 & 27.56 & 31.62 \\

& VizWiz
& 31.94 & 33.47 & 39.95 & 18.24 & 7.64 & 26.25 \\

& OK-VQA
& 37.53 & 40.23 & 19.06 & 36.47 & 14.06 & 29.47 \\

& PathVQA
& 27.55 & 37.72 & 10.86 & 23.45 & 38.02 & 27.52 \\
\bottomrule
\end{tabular*}
\caption{
Cross-domain evaluation on target queries for which the demonstrations retained by Dual-Prior Contrastive Scoring already satisfy the required 4-shot context length (Accuracy, \%).
No dynamic-pool supplementation is activated in these cases, allowing the contribution of the dual-prior scoring component to be evaluated independently.
Base denotes the pseudo-label baseline, and Avg.\ denotes the average across the five target datasets for each source domain.
Best results are in bold.
}
\label{supp:tab:dpcs_only}
\end{table*}

\begin{table*}[t]
\ContinuedFloat
\centering
\small
\setlength{\tabcolsep}{3.5pt}

\begin{tabular*}{\textwidth}{
@{\extracolsep{\fill}}
llcccccc
@{}
}
\toprule
\textbf{Method}
& \textbf{Source}
& \textbf{VQAv2}
& \textbf{Number}
& \textbf{VizWiz}
& \textbf{OK-VQA}
& \textbf{PathVQA}
& \textbf{Avg.} \\
\midrule

\multirow{5}{*}{\textbf{MMICES}}
& VQAv2
& 47.30 & 39.03 & 19.90 & 29.83 & 18.49 & 30.91 \\

& Number
& 42.86 & 42.53 & 15.51 & 25.78 & \textbf{29.38} & 31.21 \\

& VizWiz
& 35.48 & 32.96 & 40.92 & 21.94 & 10.60 & 28.38 \\

& OK-VQA
& 39.27 & 38.35 & 19.00 & 35.36 & 12.33 & 28.86 \\

& PathVQA
& 27.01 & 35.47 & 9.72 & 20.96 & 37.44 & 26.12 \\
\midrule

\multirow{5}{*}{\textbf{RICES}}
& VQAv2
& 48.03 & 38.49 & 20.73 & 30.48 & 22.65 & 32.08 \\

& Number
& 43.84 & 44.77 & 17.46 & 28.41 & 28.49 & 32.59 \\

& VizWiz
& 37.26 & 34.25 & 43.34 & \textbf{25.38} & 11.40 & 30.33 \\

& OK-VQA
& 39.78 & 40.00 & 18.66 & 37.80 & 17.41 & 30.73 \\

& PathVQA
& 27.84 & 38.23 & 11.46 & 22.11 & 36.38 & 27.20 \\
\midrule

\multirow{5}{*}{\textbf{DARA}}
& VQAv2
& 48.90 & \textbf{41.03} & 17.46 & 28.88 & 18.72 & 31.00 \\

& Number
& 43.04 & 44.59 & 12.84 & 27.05 & 26.43 & 30.79 \\

& VizWiz
& 33.79 & 41.84 & 48.73 & 22.10 & 10.84 & 31.46 \\

& OK-VQA
& 39.33 & 42.95 & 16.50 & 31.48 & 15.68 & 29.19 \\

& PathVQA
& 30.69 & \textbf{41.44} & 7.88 & 26.15 & 28.88 & 27.01 \\
\midrule

\multirow{5}{*}{\textbf{Ours}}
& VQAv2
& \textbf{50.42} & 40.03 & \textbf{20.93}
& \textbf{31.32} & \textbf{25.06} & \textbf{33.55} \\

& Number
& \textbf{47.43} & \textbf{47.73} & 18.29
& \textbf{28.45} & 29.34 & \textbf{34.25} \\

& VizWiz
& \textbf{43.05} & \textbf{45.15} & \textbf{49.80}
& 23.24 & \textbf{20.92} & \textbf{36.43} \\

& OK-VQA
& \textbf{43.30} & \textbf{44.28} & 19.00
& \textbf{38.91} & \textbf{22.68} & \textbf{33.63} \\

& PathVQA
& \textbf{38.70} & 40.38 & \textbf{12.96}
& \textbf{27.11} & \textbf{40.99} & \textbf{32.03} \\
\bottomrule

\end{tabular*}
\caption[]{Cross-domain evaluation on the DPCS-sufficient subset (continued).}
\end{table*}


\begin{table*}[t]
\centering
\small
\setlength{\tabcolsep}{2.7pt}
\newcommand{\gaincell}[3]{\shortstack{#1 $\rightarrow$ \textbf{#2}\\$(+#3)$}}

\begin{tabular*}{\textwidth}{
@{\extracolsep{\fill}}
lcccccc
@{}
}
\toprule
\textbf{Source}
& \textbf{VQAv2}
& \textbf{Number}
& \textbf{VizWiz}
& \textbf{OK-VQA}
& \textbf{PathVQA}
& \textbf{Avg.} \\
\midrule

VQAv2
& \gaincell{45.16}{49.60}{4.44}
& \gaincell{36.40}{38.97}{2.57}
& \gaincell{17.30}{19.40}{2.10}
& \gaincell{24.80}{29.83}{5.03}
& \gaincell{17.70}{23.35}{5.65}
& \gaincell{28.27}{32.23}{3.96} \\

Number
& \gaincell{34.30}{45.12}{10.82}
& \gaincell{44.43}{45.30}{0.87}
& \gaincell{12.80}{15.04}{2.24}
& \gaincell{22.70}{27.88}{5.18}
& \gaincell{20.10}{27.45}{7.35}
& \gaincell{26.87}{32.16}{5.29} \\

VizWiz
& \gaincell{29.76}{41.76}{12.00}
& \gaincell{41.60}{42.27}{0.67}
& \gaincell{46.90}{48.72}{1.82}
& \gaincell{18.10}{22.46}{4.36}
& \gaincell{17.30}{19.51}{2.21}
& \gaincell{30.73}{34.94}{4.21} \\

OK-VQA
& \gaincell{33.68}{41.66}{7.98}
& \gaincell{39.05}{42.81}{3.76}
& \gaincell{16.80}{17.96}{1.16}
& \gaincell{35.12}{35.94}{0.82}
& \gaincell{13.85}{21.15}{7.30}
& \gaincell{27.70}{31.90}{4.20} \\

PathVQA
& \gaincell{32.70}{35.87}{3.17}
& \gaincell{36.95}{38.29}{1.34}
& \gaincell{10.55}{12.21}{1.66}
& \gaincell{19.10}{23.88}{4.78}
& \gaincell{36.95}{38.10}{1.15}
& \gaincell{27.25}{29.67}{2.42} \\

\bottomrule
\end{tabular*}
\caption{
Effect of Progressive Contextual Evolution on target queries for
which Dual-Prior Contrastive Scoring retains fewer than the required
$K=4$ demonstrations (Accuracy, \%).
Results are reported as
``without supplementation $\rightarrow$ with supplementation'',
with the absolute improvement shown in parentheses.
Without supplementation, inference uses only the valid demonstrations
retained by DPCS, resulting in a variable-length context.
With supplementation, the remaining context positions are filled
through Progressive Contextual Evolution.
Avg.\ denotes the average across the five target datasets.
}
\label{supp:tab:supplement_effect}
\end{table*}

\begin{table*}[t]
\centering

\small
\setlength{\tabcolsep}{3.2pt}

\begin{tabular*}{0.72\textwidth}{@{\extracolsep{\fill}}lcccc@{}}
\toprule
\textbf{Target}
& \textbf{Admitted/Evaluated}
& \textbf{Admission}
& \textbf{Correct}
& \textbf{Admission} \\
\textbf{Dataset}
& \textbf{Samples}
& \textbf{Rate (\%)}
& \textbf{Samples}
& \textbf{Acc. (\%)} \\
\midrule
VQAv2
& 12,437 / 50,000
& 24.87
& 9,427
& 75.80 \\

Number
& 4,283 / 13,775
& 31.09
& 3,435
& 80.20 \\

VizWiz
& 4,859 / 21,595
& 22.50
& 3,329
& 68.51 \\

OK-VQA
& 5,198 / 25,230
& 20.60
& 3,701
& 71.20 \\

PathVQA
& 8,544 / 31,295
& 27.30
& 6,178
& 72.31 \\
\midrule
\textbf{Overall}
& \textbf{35,321 / 141,895}
& \textbf{24.89}
& \textbf{26,070}
& \textbf{73.81} \\
\bottomrule
\end{tabular*}
\caption{
Quality of pseudo-labeled samples admitted into the dynamic pool.
Admitted/Evaluated reports the number of accepted samples relative to
all processed target queries.
Admission rate denotes the proportion of queries accepted by dual
adaptive gating, while admission accuracy measures the proportion of
admitted pseudo-labels that match the ground-truth answers.
Results are aggregated across all five source domains.
}
\label{supp:tab:dynamic_pool_quality}
\end{table*}


\begin{table*}[t]
\centering

\small
\setlength{\tabcolsep}{4.8pt}

\begin{tabular*}{\textwidth}{
@{\extracolsep{\fill}}
llcccccc
@{}
}
\toprule
\multirow{2}{*}{\textbf{Source}}
& \multirow{2}{*}{\textbf{Method}}
& \multicolumn{5}{c}{\textbf{Target}}
& \multirow{2}{*}{\textbf{Avg.}}
\\
\cmidrule(lr){3-7}
&
& \textbf{VQAv2}
& \textbf{Number}
& \textbf{VizWiz}
& \textbf{OK-VQA}
& \textbf{PathVQA}
&
\\
\midrule

\multirow{3}{*}{\textbf{VQAv2}}
& Image sim
& 47.10 & 39.93 & 19.02 & 28.59 & 21.66 & 31.26
\\
& Question sim
& 47.51 & 36.77 & 19.44 & 28.51 & 19.76 & 30.40
\\
& QA sim
& 48.56 & 36.88 & 19.22 & 29.34 & 20.07 & 30.81
\\
\midrule

\multirow{3}{*}{\textbf{Number}}
& Image sim
& 44.42 & 46.10 & 15.34 & 27.07 & 29.73 & 32.53
\\
& Question sim
& 40.52 & 45.84 & 16.33 & 24.17 & 28.58 & 31.09
\\
& QA sim
& 42.85 & 44.83 & 16.25 & 25.79 & 27.88 & 31.52
\\
\midrule

\multirow{3}{*}{\textbf{VizWiz}}
& Image sim
& 37.30 & 35.54 & 42.56 & 24.58 & 12.51 & 30.50
\\
& Question sim
& 32.08 & 34.19 & 36.01 & 20.78 & 9.99 & 26.61
\\
& QA sim
& 32.72 & 31.94 & 40.00 & 19.47 & 9.00 & 26.63
\\
\midrule

\multirow{3}{*}{\textbf{OK-VQA}}
& Image sim
& 39.85 & 39.38 & 18.31 & 33.21 & 17.64 & 29.68
\\
& Question sim
& 35.85 & 39.46 & 19.36 & 34.78 & 14.44 & 28.78
\\
& QA sim
& 38.74 & 39.85 & 18.14 & 36.68 & 15.37 & 29.76
\\
\midrule

\multirow{3}{*}{\textbf{PathVQA}}
& Image sim
& 28.70 & 38.11 & 10.04 & 23.01 & 28.25 & 25.62
\\
& Question sim
& 25.19 & 36.19 & 8.67 & 19.66 & 36.89 & 25.32
\\
& QA sim
& 27.86 & 36.30 & 10.77 & 22.04 & 36.94 & 26.78
\\

\bottomrule
\end{tabular*}
\caption{Cross-domain evaluation of similarity-based retrieval methods
under different source datasets (Accuracy, \%).
Image sim, Question sim, and QA sim denote image-, question-,
and question--answer-based similarity retrieval, respectively.}
\label{supp:tab:detail_sim}
\end{table*}


\begin{table*}[t]
\centering

\small
\setlength{\tabcolsep}{3pt}

\begin{tabular*}{0.72\textwidth}{
@{\extracolsep{\fill}}
lccccccc
@{}
}
\toprule
\textbf{Target}
& \textbf{Base}
& \textbf{TT-RAA}
& \textbf{BCA}
& \textbf{MMICES}
& \textbf{RICES}
& \textbf{DARA}
& \textbf{Ours}
\\
\midrule

\textbf{VQAv2}
& 42.22
& 27.28
& 42.85
& 42.44
& 44.42
& 43.87
& \textbf{45.61}
\\

\textbf{Number}
& 45.93
& 36.19
& 44.83
& 43.70
& \textbf{46.10}
& 45.08
& 46.05
\\

\textbf{VizWiz}
& \textbf{18.10}
& 12.64
& 16.18
& 15.71
& 16.33
& 12.58
& 14.48
\\

\textbf{OK-VQA}
& 24.40
& 17.12
& 25.79
& 25.92
& 27.07
& 27.86
& \textbf{28.20}
\\

\textbf{PathVQA}
& 27.02
& 24.73
& 27.86
& 28.02
& \textbf{29.73}
& 27.27
& 25.92
\\

\midrule

\textbf{Avg.}
& 31.53
& 23.59
& 31.50
& 31.16
& \textbf{32.73}
& 31.33
& 32.05
\\
\\

\bottomrule
\end{tabular*}
\caption{Cross-domain evaluation with Number as the source domain
(4-shot accuracy, \%). Best results are in bold.}
\label{supp:tab:number_results}
\end{table*}

\begin{table*}[t]
\centering
\footnotesize

\begin{tabular*}{0.95\textwidth}{@{\extracolsep{\fill}} ccccccc @{}}
\toprule
\textbf{Source Domain} & \textbf{VQAv2} & \textbf{Number} & \textbf{VizWiz} & \textbf{OK-VQA} & \textbf{PathVQA} & \textbf{Avg.} \\
\midrule
\textbf{VQAv2}   & 52.37 & 43.19 & 25.76 & 37.02 & 20.84 & 35.84 \\
\textbf{Number}  & 48.58 & 45.00 & 24.25 & 33.41 & 35.57 & 37.36 \\
\textbf{VizWiz}  & 48.32 & 43.62 & 49.08 & 29.88 & 15.78 & 37.34 \\
\textbf{OK-VQA}  & 51.37 & 48.50 & 28.08 & 40.64 & 28.45 & 39.41 \\
\textbf{PathVQA} & 37.11 & 44.92 & 18.72 & 32.91 & 42.08 & 35.15 \\
\midrule
\multicolumn{7}{c}{\normalsize (a) Performance with Qwen2-VL} \\
\midrule
\textbf{Source Domain} & \textbf{VQAv2} & \textbf{Number} & \textbf{VizWiz} & \textbf{OK-VQA} & \textbf{PathVQA} & \textbf{Avg.} \\
\midrule
\textbf{VQAv2}   & 50.63 & 41.57 & 24.71 & 30.49 & 21.03 & 33.69 \\
\textbf{Number}  & 47.25 & 45.13 & 20.50 & 34.58 & 22.65 & 34.02 \\
\textbf{VizWiz}  & 47.43 & 38.27 & 48.17 & 26.04 & 14.81 & 34.94 \\
\textbf{OK-VQA}  & 48.68 & 44.90 & 27.64 & 42.12 & 26.33 & 37.93 \\
\textbf{PathVQA} & 35.78 & 45.72 & 14.62 & 31.69 & 40.70 & 33.70 \\
\midrule
\multicolumn{7}{c}{\normalsize (b) Performance with Llama-3-V} \\
\bottomrule
\end{tabular*}
\caption{Cross-domain evaluation of our proposed PACE framework across diverse model backbones (Accuracy, \%). All reported metrics represent the absolute performance of our method. (a) Performance utilizing the Qwen2-VL architecture. (b) Performance utilizing the Llama-3-V architecture. The results demonstrate that our framework consistently mitigates severe distribution shifts regardless of the underlying base model, highlighting its architecture-agnostic robustness.}
\label{supp:tab:model}
\label{supp:tab:sub_a}
\label{supp:tab:sub_b}
\end{table*}

\begin{table*}[t]
\centering
\footnotesize

\begin{tabular*}{0.95\textwidth}{@{\extracolsep{\fill}} ccccccc @{}}
\toprule
\textbf{Source Domain} & \textbf{VQAv2} & \textbf{Number} & \textbf{VizWiz} & \textbf{OK-VQA} & \textbf{PathVQA} & \textbf{Avg.} \\
\midrule

\textbf{VQAv2}   & 40.38 & 30.81 & 15.17 & 21.88 & 6.43  & 22.93 \\
\textbf{Number}  & 27.65 & 37.03 & 9.97  & 14.86 & 14.07 & 20.72 \\
\textbf{VizWiz}  & 21.66 & 23.00 & 35.94 & 11.98 & 5.04  & 19.52 \\
\textbf{OK-VQA}  & 25.17 & 26.33 & 12.19 & 30.86 & 11.49 & 21.21 \\
\textbf{PathVQA} & 19.92 & 22.71 & 8.58  & 12.39 & 33.14 & 19.35 \\

\bottomrule
\end{tabular*}
\caption{Cross-domain evaluation of PACE initialized with zero-shot
pseudo-labels (Accuracy, \%). All values report the absolute
performance of our method when the initial target-domain
pseudo-labels are generated using zero-shot inference instead of
the default source-domain ICL strategy. Compared with standard
PACE, zero-shot initialization leads to a substantial performance
degradation, indicating that the structural priors provided by
source-domain ICL are important for subsequent contextual evolution.}
\label{supp:tab:zs}
\end{table*}

\section{Datasets}

To rigorously evaluate the cross-domain generalization capability of the proposed Progressive Arbitration of Contextual Evidence framework, we constructed a comprehensive benchmark utilizing five distinct Visual Question Answering (VQA) datasets. These datasets were strategically selected to represent diverse and severe distribution shifts in visual appearance, semantics, and task objectives. The detailed characteristics of each dataset are described below:


\textbf{VQAv2 \cite{Goyal_2017_CVPR}.} Serving as our standard natural-image baseline, VQAv2 comprises diverse, high-quality real-world images sourced from the MS COCO dataset, paired with open-ended questions. The dataset demands comprehensive visual understanding, encompassing a wide range of question types conventionally categorized into ``yes/no'' (boolean logic), ``number'' (counting), and ``other'' (e.g., attribute recognition, spatial relationships, and object identification). When VQAv2 is utilized as the source domain, we adopted its entire official training split—consisting of 443,757 questions based on 82,783 images—to construct the support pool ($\mathcal{S}_l$). For target-domain evaluation, since the ground-truth annotations for the official test split are withheld for online leaderboard evaluation, we utilized the validation split, which originally contains 214,354 questions associated with 40,504 images. To maintain scale parity across our evaluation benchmark and prevent this exceptionally large dataset from dominating the overall performance metrics, we uniformly sampled exactly 10,000 queries from this validation split to form the target-domain inference set ($\mathcal{S}_u$).

\textbf{OK-VQA \cite{marino2019ok}.} Distinct from conventional visual understanding, OK-VQA requires the model to incorporate external knowledge, commonsense reasoning, or factual information not explicitly present in the image to answer the queries. The integration of this dataset introduces a severe \textbf{semantic shift}, rigorously testing whether the model degenerates into superficial visual matching when underlying semantic reasoning is strictly required. \textit{For instance, a query might show a fire hydrant and ask for the century it was first patented, or present a specific brand of soda and inquire about the company that produces it.} Such questions force the model to look beyond visual pixels and retrieve external knowledge anchors. For our experiments, we utilized its official standard splits: the entire training set comprising 9,009 questions was employed as the source domain ($\mathcal{S}_l$), while the complete test set, containing 5,046 questions, functioned as the target-domain inference set ($\mathcal{S}_u$).

\textbf{VizWiz \cite{Gurari_2018_CVPR}.} Originating from photographs taken by visually impaired individuals navigating everyday settings, this dataset introduces a drastic \textbf{visual quality shift}. Unlike the pristine, well-framed images typical of standard benchmarks, VizWiz images frequently suffer from severe degradation, including motion blur, extreme over/under-exposure, and poor framing where the subject of interest is often partially or entirely occluded. \textit{For instance, a query might feature a heavily blurred, off-center photo of a generic bottle and ask, ``What kind of medication is this?'', or present a completely dark image and inquire, ``What color is my shirt?''} Such examples frequently render the questions inherently ``unanswerable,'' challenging the model to recognize the limits of available visual evidence rather than hallucinating text-driven guesses. Following the official VQA task partitions, we utilized the entire training set of 20,523 questions as the source domain ($\mathcal{S}_l$), and the complete validation set comprising 4,319 questions as the target-domain inference set ($\mathcal{S}_u$).

\textbf{PathVQA \cite{he2020pathvqa}.} Representing an \textbf{extreme domain shift}, this dataset departs entirely from the everyday natural scenes prevalent in standard pre-training. PathVQA consists exclusively of medical pathology slides, demanding highly specialized histological knowledge and structural visual interpretation. \textit{For instance, a query might present a microscopic H\&E stained tissue patch and ask, ``Does this slide show evidence of necrosis?'' or ``Are there any malignant cells present?''} The integration of PathVQA critically evaluates whether the framework can adapt to a completely foreign visual modality without suffering from catastrophic failure. For our experimental pipeline, we utilized the standard training split of 19,654 questions as the source domain ($\mathcal{S}_l$), and the complete test split comprising 6,259 questions as the target-domain inference set ($\mathcal{S}_u$).

\textbf{Number \cite{ren2015exploring}.} Unlike the previously introduced datasets that present visual or semantic domain gaps, this dataset focuses strictly on evaluating a \textbf{task-specific shift}. It is a highly specialized subset rigorously filtered from the broader COCO-QA benchmark, isolating only the queries that strictly necessitate object counting and spatial enumeration. \textit{For example, rather than general descriptions, queries strictly follow formats like, ``How many red apples are on the tree?'' or ``How many zebras are standing in the grass?''} By constricting the evaluation to numerical counting, we aim to assess whether the model can disentangle the core spatial-reasoning mechanism from the statistical language priors of general VQA tasks. We utilized the standard COCO-QA Number training split containing 5,885 questions as the source domain ($\mathcal{S}_l$), and its corresponding test split of 2,755 questions as the target-domain inference set ($\mathcal{S}_u$).

\section{Implementation}

To ensure the reproducibility of our study, we expand upon the foundational settings provided in the main paper by detailing the underlying tensor-level operations, specific multimodal prompting formats, and evaluation protocols.

\noindent\textbf{Multimodal Prompt Template and Input Packing.} Across all experiments, we strictly aligned our input formatting with the OpenFlamingo architecture \cite{awadalla2023openflamingo}. For a given context sequence, each demonstration sample was assembled by interleaving the image tensor with the following text template:
\begin{center}
\begin{minipage}{0.95\linewidth}
\centering
\small
\texttt{\detokenize{<image> Question: {q} Answer: {a} <|endofchunk|>}}
\end{minipage}
\end{center}
During the dynamic context evolution (Phase 2), if a target-domain sample with a pseudo-label was selected for the context, its pseudo-label directly replaced the ground-truth answer in the \texttt{\{a\}} placeholder to maintain structural consistency. For the target query, the sequence was appended without the termination token: \texttt{<image> Question: \{q\_t\} Answer:}, which safely prevented any answer leakage and triggered the autoregressive generation. On the tensor level, all images were processed by the standard visual processor, stacked, expanded via double \texttt{unsqueeze} operations, and cast to half-precision (\texttt{FP16}) before being fed to the visual encoder.

\noindent\textbf{Tensor-level Implementation of Visual Masking.} In the calculation of the visual information deprivation penalty ($f_m$), we did not substitute the visual features with a \texttt{torch.zeros()} tensor, as this would violate the expected input distribution of the visual encoder. Instead, the masking was implemented by generating a purely black PIL image ($224 \times 224$, RGB=($0,0,0$)), which was subsequently passed through the exact same \texttt{vis\_processor} pipeline (incorporating standard resizing and pixel normalization) as the normal images. This normalized ``blindfold'' tensor replaced only the query image $v_t$, while all images within the contextual demonstrations remained intact and visible.

\noindent\textbf{Text Normalization and Matching Protocols.} To determine logical consistency and compute final accuracies, we implemented a rigorous text normalization pipeline strictly aligned with the official VQA evaluation logic. Before matching, both the generated sequence and the reference labels underwent: (1) conversion to lowercase; (2) punctuation removal (utilizing regular expressions to eliminate non-decimal periods); (3) contraction restoration (e.g., mapping ``dont'' to ``don't''); (4) word-to-digit conversion for English numbers (e.g., ``two'' to ``2''); and (5) removal of standard articles and excess whitespace.

For the scoring mechanism, datasets providing multiple ground-truth annotations per query (VQAv2, OK-VQA, VizWiz) were evaluated using the standard robust metric: $\min(\text{count}/3, 1.0)$, where ``count'' is the number of exact matches with annotators \cite{Antol_2015_ICCV}. Conversely, for single-annotation datasets (PathVQA, Number), strict 0/1 exact matching was applied.

\noindent\textbf{Hardware and Computational Resources.} All offline inference procedures, feature extraction for the various retrieval baselines (e.g., Image-Sim, QA-Sim), and the comprehensive contrastive reranking stages were executed on a computing cluster equipped with 8 NVIDIA A800 GPUs. While our Progressive Arbitration of Contextual Evidence framework is entirely training-free, the exhaustive all-to-all cross-domain evaluation matrix—combined with the multiple forward passes required for the Dual-Prior Contrastive Scoring across different similarity baselines—necessitated extensive computation. Completing the full suite of experiments required approximately one week of wall-clock time on this 8-GPU cluster.

\section{Extended Experiments}

In this section, we provide extended results to further investigate the performance boundaries of the PACE framework across various dimensions, including sensitivity to context length, model architecture, and initialization strategies.

\noindent\textbf{Analysis of Varying Few-shot Sizes.} We evaluated the performance evolution of our framework under the settings of $K \in \{1, 2, 4, 8\}$. Experimental results indicate a consistent upward trend in average accuracy as $K$ increases (see Table \ref{supp:tab:shot}). Notably, in extremely low-shot scenarios (e.g., $K=1$ or $2$), the number of valid candidates $\mathcal{E}_{valid}$ filtered via the Dual-Prior Contrastive Scoring (DPCS) often suffices to fill the limited contextual quota, resulting in a lower activation frequency of the dynamic pool $\mathcal{S}_{dyn}$. This phenomenon effectively serves as an implicit ablation study, demonstrating the standalone precision of the DPCS mechanism: it successfully identifies instances with genuine reasoning gains even without the benefit of long-term test-stream accumulation. As $K$ scales to $8$, the continued performance growth highlights the framework's scalability and its ability to leverage deeper target-domain insights from the dynamic pool while safely mitigating potential noise accumulation from extended context windows.

\noindent\textbf{Isolating Dual-Prior Contrastive Scoring.}
To further disentangle the contribution of Dual-Prior Contrastive
Scoring (DPCS) from Progressive Contextual Evolution, we conduct
a controlled subset analysis on target queries for which the
logit-based filtering stage already retains at least the required
number of demonstrations. Specifically, for the 4-shot setting,
we define the diagnostic subset as
\[
\mathcal{Q}_{\mathrm{DPCS}}
=
\left\{
x_t \mid
\left|\mathcal{E}_{\mathrm{valid}}(x_t)\right|
\geq 4
\right\}.
\]
For every query in this subset, the contextual gap satisfies
$\Delta K=0$, and therefore no supplementary demonstrations are
drawn from either the dynamic target-domain pool or the labeled
source-domain pool. The final context is constructed entirely
from the highest-ranked candidates retained by DPCS. Consequently,
the result of our method on this subset reflects the effect of the
dual-prior scoring component without intervention from progressive
target-domain accumulation. For a controlled comparison, all
baselines are evaluated on exactly the same target-query subset.

As shown in Table~\ref{supp:tab:dpcs_only}, the DPCS-only configuration
achieves the highest source-averaged accuracy under all five source
domains, obtaining 33.55\%, 34.25\%, 36.43\%, 33.63\%, and 32.03\%
with VQAv2, Number, VizWiz, OK-VQA, and PathVQA as the source
domain, respectively. Averaged across the five source domains,
our method reaches 33.98\%, outperforming the strongest overall
baseline, RICES, by 3.39 percentage points
(33.98\% vs.\ 30.59\%). The gains are particularly pronounced
with VizWiz and PathVQA as the source domains, where our method
improves over the strongest competing approach by 4.97 and
4.51 points, respectively.

These results indicate that the performance gain of our framework
does not arise solely from the progressive incorporation of
pseudo-labeled target-domain samples. Even when the dynamic pool
is never activated, the dual-prior margin can identify
demonstrations whose predictive contribution extends beyond both
target-image-independent shortcuts and generic source-context
biases. This provides direct evidence that DPCS constitutes an
effective standalone selection mechanism, while Progressive
Contextual Evolution serves as a complementary mechanism for
queries whose valid candidates are insufficient to fill the
required context length.

\noindent\textbf{Effect of Progressive Contextual Evolution.}
We further examine the complementary role of Progressive Contextual
Evolution on the subset of target queries for which DPCS retains fewer
than the required $K=4$ valid demonstrations. For these queries, directly
performing inference with the incomplete context leads to substantially
lower accuracy, as the limited number of retained demonstrations cannot
provide sufficient task and answer priors. We therefore compare inference
using only the valid DPCS demonstrations with inference after filling the
remaining context positions through progressive supplementation.

As shown in Table~\ref{supp:tab:supplement_effect}, contextual supplementation
consistently improves performance across all five source domains. The
source-domain average accuracy increases from 28.27\% to 32.23\% with
VQAv2 as the source, from 26.87\% to 32.16\% with Number, from 30.73\%
to 34.94\% with VizWiz, from 27.70\% to 31.90\% with OK-VQA, and from
27.25\% to 29.67\% with PathVQA. Averaged across the five source domains,
Progressive Contextual Evolution improves accuracy from 28.16\% to
32.18\%, corresponding to an average gain of 4.02 percentage points.
These results show that leaving the contextual slots unfilled causes a
clear performance degradation, whereas progressive supplementation
recovers a substantial portion of the lost performance by introducing
additional target-aligned evidence.

Nevertheless, the supplemented subset still performs below the subset
for which DPCS alone already satisfies the required context length.
Specifically, its source-domain averages remain 1.32, 2.09, 1.49, 1.73,
and 2.36 percentage points lower than the corresponding DPCS-sufficient
results reported in Table~\ref{supp:tab:dpcs_only}. This gap is expected:
queries that retain at least four valid demonstrations possess
a more reliable and complete candidate context, whereas queries requiring
supplementation constitute a more difficult subset with insufficient
high-confidence candidates. Progressive Contextual Evolution therefore
does not completely eliminate the intrinsic difficulty of these queries,
but it substantially narrows the performance gap caused by an incomplete
context.

Together, the two subset analyses clarify the complementary roles of the
proposed components. DPCS provides a strong standalone selection mechanism
when sufficient valid demonstrations are available, while Progressive
Contextual Evolution serves as a corrective mechanism when the filtered
context is insufficient, recovering most of the resulting performance
loss without overriding the high-quality evidence retained by DPCS.

\noindent\textbf{Quality of Dynamic-Pool Admissions.}
We further assess the reliability of pseudo-labeled samples admitted
into the dynamic pool using ground-truth annotations. This evaluation
is performed only for analysis and does not expose target labels during
inference. As shown in Table~\ref{supp:tab:dynamic_pool_quality}, dual
adaptive gating admits approximately 24.89\% of the processed target
queries, among which 73.81\% receive correct pseudo-labels. Admission
accuracy remains relatively stable across heterogeneous target domains,
with stronger reliability on the structurally constrained Number
dataset and lower accuracy on the visually challenging VizWiz dataset.
These results indicate that the agreement and confidence gates filter
out a substantial proportion of unreliable predictions, limiting noise
accumulation during progressive contextual evolution.

\noindent\textbf{Detailed Analysis of Similarity-Based Retrieval.}
In the main paper, we report RICES as a consolidated similarity-based baseline, where, for each source--target pair, its result is obtained by selecting the best-performing variant among image similarity, question similarity, and question--answer similarity. To provide a more transparent comparison, Table~\ref{supp:tab:detail_sim} presents the complete results of these three retrieval strategies separately across all source and target datasets. The results show that no single similarity metric consistently dominates across all domain combinations. Image similarity generally provides stronger average performance for several source domains, whereas question or question--answer similarity can be more effective for particular target datasets. This variation indicates that the optimal retrieval signal is highly dependent on the specific source--target distribution shift. Therefore, the consolidated RICES results reported in the main paper represent the strongest achievable performance of conventional single-similarity retrieval, providing a competitive reference for evaluating our framework.

\begin{figure*}[t]
    \centering
    \includegraphics[width=\textwidth]{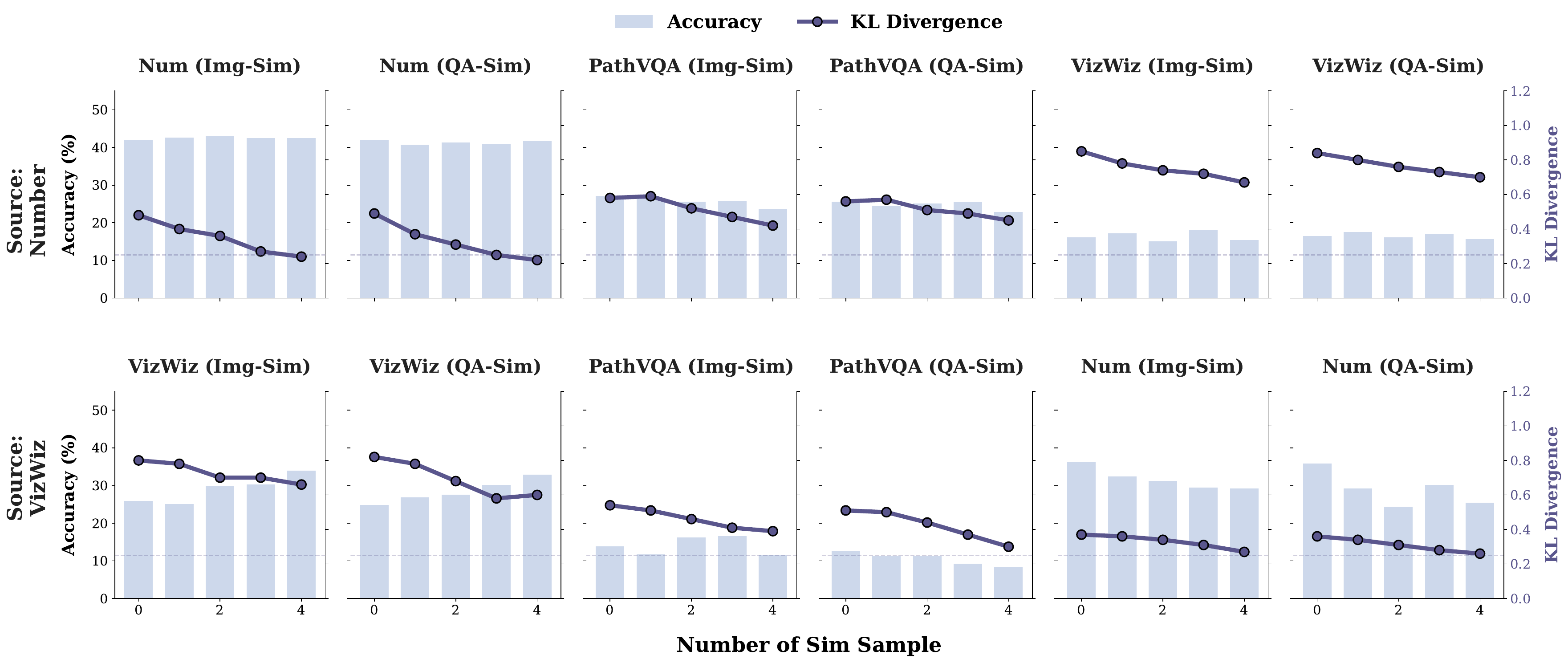}
    \caption{
    Complete analysis of the impact of similarity-retrieved demonstrations on
    accuracy and visual reliance. Bars report accuracy, while curves report
    $D_{\mathrm{KL}}$ as the number of similarity-retrieved samples increases.
    The full results cover image-similarity and QA-similarity retrieval across
    all evaluated source--target combinations. Lower $D_{\mathrm{KL}}$ indicates
    reduced sensitivity to the target image.
    }
    \label{supp:fig:kl_full}
\end{figure*}

\begin{figure}[t]
    \centering
    \includegraphics[width=\columnwidth]{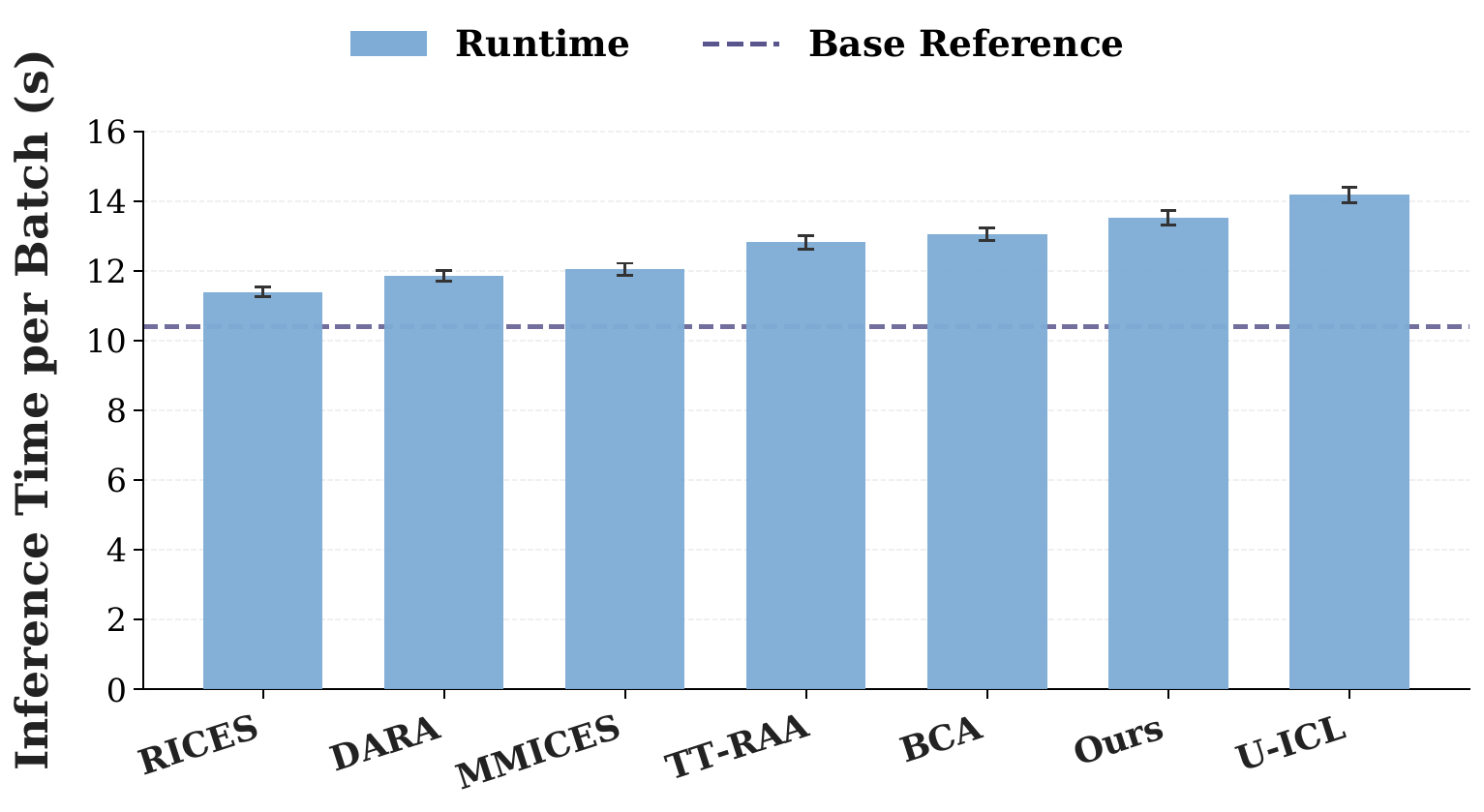}
    \caption{
    Average online inference time per batch under the 4-shot setting.
    Error bars denote standard deviations, and the dashed line indicates
    the runtime of Base.
    }
    \label{supp:fig:runtime}
\end{figure}

\noindent\textbf{Results with Number as the Source Domain.}
Due to space and layout constraints, the main paper reports
results for four representative source domains, while the complete
evaluation with Number as the source domain is provided in
Table~\ref{supp:tab:number_results}. Compared with the other source
datasets, Number has a relatively narrow and highly structured
answer space, which may limit the transferability of its
demonstrations to visually and semantically diverse target domains.
Our framework achieves an average accuracy of 32.05\%, remaining
competitive with the strongest baseline, RICES, at 32.73\%.
Notably, it obtains the best performance on VQAv2 and OK-VQA,
outperforming the strongest competing method by 1.19 and 0.34
percentage points, respectively, and trails RICES by only 0.05
points on the Number target. These results indicate that the
proposed contextual evolution strategy remains effective even
when initialized from a structurally specialized source domain.

\noindent\textbf{Complete Analysis of Similarity-Induced Visual Reliance.}
Due to space constraints, the main paper presents eight representative panels of the similarity-retrieval analysis. Figure~\ref{supp:fig:kl_full} provides the complete results across all twelve source--target and retrieval-strategy combinations. The additional panels exhibit the same overall tendency: as the number of similarity-retrieved demonstrations increases, $D_{\mathrm{KL}}$ generally decreases, indicating weakened dependence on the target image. However, the corresponding effect on accuracy depends strongly on domain alignment. In the in-domain VizWiz $\rightarrow$ VizWiz setting, accuracy remains relatively stable or even improves slightly despite the reduction in $D_{\mathrm{KL}}$, suggesting that aligned textual regularities can partially compensate for reduced visual reliance. In contrast, under the additional cross-domain settings VizWiz $\rightarrow$ Number and VizWiz $\rightarrow$ PathVQA, the decline in $D_{\mathrm{KL}}$ is accompanied by an overall accuracy degradation, particularly under QA-similarity retrieval. These complete results further confirm that similarity retrieval introduces semantic redundancy that suppresses visual reliance, while its downstream impact is substantially more harmful when the retrieved source-domain patterns are misaligned with the target domain.

\noindent\textbf{Robustness across Diverse Model Backbones.} To verify the architecture-agnostic nature of PACE, we extended our evaluation from the OpenFlamingo baseline to more advanced large vision-language models, including the Qwen2-VL \cite{wang2024qwen2} and Llama-3-V series \cite{grattafiori2024llama}. As shown in Table \ref{supp:tab:model}, our framework maintains highly robust absolute performance across different visual encoders (e.g., CLIP \cite{radford2021learning}, SigLIP \cite{zhai2023sigmoid}) and language decoder architectures, achieving significant performance improvements across all target datasets. Rather than merely comparing to baselines, these results underscore that the framework consistently neutralizes severe distribution shifts regardless of the underlying base model. This demonstrates that the inherent biases addressed by our framework are universal challenges for LVLMs in cross-domain scenarios, and that our proposed method acts as a highly adaptive, plug-and-play solution.

\noindent\textbf{Impact of Pseudo-Label Generation Strategy.}
To examine how pseudo-label generation affects downstream
performance, we replace the default source-domain ICL strategy
with zero-shot inference when generating the initial target-domain
pseudo-labels. All subsequent components of PACE, including
DPCS, contextual supplementation, and dual adaptive gating, remain
unchanged. As shown in Table~\ref{supp:tab:zs}, zero-shot initialization
leads to a substantial performance degradation across the
source--target evaluation matrix.

These results indicate that the initial pseudo-labels produced
under a labeled source-domain ICL context provide a more reliable
foundation for subsequent contextual evolution than zero-shot
predictions. Although ICL-based pseudo-labels may inherit
source-domain biases, they generally provide stronger task and
answer priors, whereas zero-shot predictions are more susceptible
to uncertainty and output-format inconsistencies, particularly in
specialized domains such as PathVQA. The source-context bias
introduced during initialization is subsequently mitigated by the
$f_{\mathrm{anc}}$ penalty in DPCS, which suppresses candidates
whose predictive contribution is dominated by generic
source-context tendencies.

\noindent\textbf{Runtime Analysis.}
We evaluate the online inference efficiency of all methods under the
4-shot setting using a single NVIDIA A800 GPU with a batch size of 8.
All methods use precomputed retrieval features and identical decoding
configurations, and the results are averaged over three runs.

As shown in Figure~\ref{supp:fig:runtime}, our method requires approximately
$1.30\times$ the runtime of Base. Although dual-prior scoring introduces
additional forward passes, its runtime remains comparable to other
competitive methods, ranging from approximately $0.95\times$ to
$1.19\times$ their inference time. In particular, it is comparable to
TT-RAA and BCA and remains faster than U-ICL. Progressive contextual
evolution adds limited overhead because it only involves dynamic-pool
retrieval and context construction. These results demonstrate that the
accuracy gains of PACE are achieved with a moderate increase in online
inference cost.

\end{document}